\pdfoutput=1

\documentclass[11pt]{article}

\usepackage{authblk}
\usepackage{acl}

\usepackage{times}
\usepackage{latexsym}

\usepackage[T1]{fontenc}

\usepackage[utf8]{inputenc}
\usepackage{enumitem}

\usepackage{microtype}

\usepackage{inconsolata}

\usepackage{graphicx}

\title{The Non-Determinism of Small LLMs: Evidence of Low Answer Consistency in Repetition Trials of Standard Multiple-Choice Benchmarks}

\author[1]{Claudio Pinhanez}
\author[1]{Paulo Cavalin}
\author[1]{Cassia Sanctos}
\author[1]{Marcelo Grave}
\author[1]{Yago Primerano}

\affil[1]{IBM Research Brazil}

\usepackage{bibentry}

\begin{document}

\maketitle

\begin{abstract}
  This work explores the consistency of small LLMs (2B-8B parameters) in answering multiple times the same question. We present a study on known, open-source LLMs responding to 10 repetitions of questions from the multiple-choice benchmarks MMLU-Redux and MedQA, considering  different inference temperatures, small  vs. medium models (50B-80B), finetuned vs. base models, and other parameters.  We also look into the effects of requiring multi-trial answer consistency on accuracy and the trade-offs involved in deciding which model best provides both of them. To support those studies, we propose some new analytical and graphical tools. Results show that the number of questions which can be answered consistently vary considerably among models but are typically in the 50\%-80\% range for small models at low inference temperatures. Also, accuracy among consistent answers seems to reasonably correlate with overall accuracy. Results for medium-sized models seem to indicate much higher levels of answer consistency.
\end{abstract}

\section{Introduction}


Most \textit{Transformer}-based \textit{Large Language Models (LLMs)} are non-deterministic, mostly as a consequence of some random drawing processes which are commonly used inside them. This paper focuses on quantifying the non-determinism of small LLMs (2~to 8~billion parameters) by examining their consistency when answering multiple-choice questions, using standard benchmarks of both general knowledge, \textit{MMLM-Redux}~\cite{gema2025mmlu}), and medical expertise, \textit{MedQA}~\cite{jin2021disease}. In the studies described in this paper with commonly-used, open-source small LLMs, they rarely displayed anything close to determinism or high levels of answer consistency. 

Interestingly, the overall results, considering all the questions, of the accuracy in 10 repetitions of the same benchmarks for a given model are remarkably similar, but models with almost identical overall accuracy showed very distinct levels of consistency when answering a given question. We found the ability to produce consistent answers in the small models we studied being in the range of 50\%-80\%, when running inference at low temperatures. Somewhat surprisingly, we also saw accuracy levels of the consistently answered questions correlating well with the overall accuracy of the models in spite of the high-levels of non-determinism.

Answer consistency is often a requirement of most non-creative applications of LLMs, but it is often wrongly assumed to naturally happen by developers. Consider, for example, the common application of LLMs in chatbots for customer service. In many countries, answering differently to customers with identical questions is considered deception and/or discrimination and may entitle legal reparations. Similarly, patients and physicians would be very suspicious of an AI radiologist which produced contradictory readings and diagnoses of the exactly same X-ray.

Answer consistency is also an important feature when considering \textit{value alignment} and \textit{safety} of LLMs. In many situations, it is not only required that the model almost never produces an undesirable output, but also, for some questions, that it never produces a particular answer. Using improper language once in a while may be tolerable, but just one case of suggesting someone to terminate his or her life is likely to be out of question.

Recent work has explored some aspects of answer consistency but mostly in the context of non-multiple choice answers and using closed commercial LLMs often with fixed inference temperatures~\cite{atil2025nondeterminismdeterministicllmsettings,nalbandyan2025score,patwardhan2024automated,lee2024evaluating,ouyang2025empirical}. Unlike those, this work focuses on multiple-choice contexts using only open source models and varying inference temperatures. We look into the issue of inference temperature and how it affects the accuracy of the consistently-produced answers and the use of multiple-choice benchmarks to avoid the noise created by measuring similarity between answers to the same question. Also, the open source models can be safely run by ourselves, since commercial access to LLMs may include undisclosed cache and retrieval systems which may impair the evaluation of consistency.



We start by defining \textit{answer consistency} as oracle guessing at the same level and show how it is translated into an objective criteria in the case of repeated multiple-choice evaluations. 
In this process, we propose a simple but inefficient method to assure the consistency in answers where we make the LLM decline to answer questions in which it is not able to be consistent. 
We then propose a representation for consistency and an associated visual analysis tool, called the consistency plot. 

With those tools, we empirically explore answer consistency at different temperatures in 23~small models of about 8~billion parameters (2B-8B) and in 3~medium models in the range of 70~billion parameters (50B-80B), including finetuned vs. base models. 
We performed experimental studies with the MMLU-Redux benchmarks and MedQA at 3 different temperatures. 


The key contributions of this paper are:
\begin{itemize}
    \item The \textbf{definition of answer consistency at a level $c$} as equivalent to oracle guessing at the same level and its derivation to 
    the context of multiple-choice evaluation.
    \item \textbf{Experimental studies with the MMLU-Redux and MedQA benchmarks} across 26~different small and medium models, with and without finetuning, showing that \textbf{small models produce consistent answers in the range of about 50\%~to 80\%}, often at low temperatures, and with the accuracy of the consistent answers correlating to the average benchmark accuracy.
    \item Evidence from the results with 3~medium-sized models (50B-80B) that \textbf{limited consistency is mostly an issue of small models (2B-8B)}.
    \item Evidence that in the general knowledge benchmark MMLU-Redux, \textbf{increasing temperature yields a higher degree of accuracy among answers which are consistently given}.
\end{itemize}

\section{Related Work}

There is very limited work on answer consistency in the context of repetition of questions. In \cite{nalbandyan2025score},  an evaluation score is proposed with one component where the same question was asked using different seeds but only reported the average accuracy. Conversely, \cite{patwardhan2024automated} looked into consistency over identical and semantically similar prompts with non-multiple choice cybersecurity benchmarks. 
Likewise, \cite{lee2024evaluating} evaluated commercial LLMs consistency to follow instructions using common benchmarks, performing 5~repetitions at a high 1.0 temperature and found limited consistency~\cite{jiang2023mistral7b}.
Also, \cite{atil2025nondeterminismdeterministicllmsettings} looked into commercial models and explored consistency in both zero- and few-shot scenarios. 
Finally, \cite{song2024good} focused on the differences between greed and sampling decoding. 

Our work differs from those because we propose more than one component when evaluating consistency, report accuracy along with its standard deviation, look at more than one temperature, and do not use closed, commercial LLMs to avoid unknown mechanisms like caching systems. 

\textit{Multiple-choice evaluation} is often used to evaluate LLMs~\cite{singhal2023expertlevelmedicalquestionanswering,jiang2023mistral7b,nori2023generalistfoundationmodelsoutcompete,dubey2024llama3herdmodels} since it can be more objectively evaluated as opposed to open questions. This is the primary reason we used multiple-choice contexts in this work. 

A related area of work has explored sensitivity of LLMs according to changes in the input and, in particular, on simple variations on the multiple-choice answers, such as reordering. In \cite{mirzadeh2024gsmsymbolicunderstandinglimitationsmathematical}, it was shown that LLMs are negatively impacted by even small changes in the input question such as when only having choice numbers modified in math-related questions. The work described in \cite{ackerman-etal-2024-novel} proposed a metric for LLM robustness to input changes and reported important impacts on the accuracy.

Another line of research has focused on investigating the \emph{consistency} of LLMs in providing a response when the question is kept intact but with variations in other factors, such as the set of choices and parameters of the inference algorithm. An investigation on the sensitivity of choice order was reported in \cite{li-etal-2024-multiple} according to different values for the temperature parameter. In \cite{wei2024rethinkinggenerativelargelanguage}, the authors compared the results of multiple-choice evaluations and open-ended answers and found low consistency between these two methods. In \cite{pezeshkpour-hruschka-2024-large}, the \textit{MV} metric based on majority voting was proposed, in what could be considered as a simplified, non-generic version of the methods proposed here, 
a theme also explored in~\cite{wang2024answersreviewingrationalitymultiple}.

\section{Defining Answer Consistency Metric}

Most LLMs use a \textit{softmax} function at the top of the decoder stack of a Transformer-based system~\cite{vaswani2017attention} which selects the next token to be generated, considering the probabilities of each token as computed by the preceding layers as part of a weighted random choice process controlled by the \textit{temperature} of the inference process. When the temperature is zero, the most likely token is, at least theoretically, produced and therefore the generation of the output token in each cycle of the inference process is, or should be, deterministic. However, when the temperature is greater than zero the random weighted drawing of the next token makes the process non-deterministic.

Notice that in the context of this study, it is trivial to determine whether the same answer was produced from the same input and whether the answer is correct or not, since we used \textit{multi-choice benchmarks}, where an answer is just one letter identifying one of the answers 
provided by the question.

\subsection{Consistency and Oracle Guessing}\label{subsec:oracle}


A way to characterize how consistent a model is in answering a set of questions is to compare it to an \textit{oracle machine} which answers questions at a certain $c$ rate of success. Oracles were introduced by Alan Turing in 1939~\cite{turing1939systems} and are a cornerstone of complexity theory, leading to:

\textit{\textbf{Definition:} given a question, an LLM-model has} $c$-answer consistency \textit{when the model is equivalent to an oracle guessing the question at a $c$ level of correctness.}

In this work, we explore systems which exhibit $c$-answer consistency in $M$ repetitions of a set with $Q$ multiple-choice questions of $k$ choices. 
We start by considering a single multiple-choice question $q$ of $k$ choices which is repeatedly evaluated by a model $M$ times, yielding answers $llm(q_i), 1 \le i \le M$, where $llm(q_i)=1$ if, and only if, the answer is correct. In $M$ repetitions, the number of possible arrangement of choices where exactly $p$ are correct, $C_M(p)$ is, trivially:
\begin{equation}
    C_M(p) = \left( \begin{array}{c} M \\ p \end{array} \right) = \frac{M!}{(M-p)!\,p!}
\end{equation}

Now consider a system with a success guessing rate $r$, where $r=1/k$ if it were a purely random guess. It is easy to see that the probability of guessing correctly exactly $p$ of the $M$ repetitions, $T_r^M(p)$ is, by basic probability:

\begin{equation} 
    P(T_r^M(p)) = C_M(p)r^p(1-r)^{(M-p)}
\end{equation}

Following, the probability of guessing correctly $p$ or more answers in $M$ repetitions, $\overline{T}_r^M(p)$ is, just by summing up:
\begin{equation}\label{eq:guessing-repetition}
    P(\overline{T}_r^M(p)) = \sum_{j=p}^{M}{C_M(j)r^j(1-r)^{(M-j)}}
\end{equation}
For instance, let us consider the value of $\overline{T}_r^M(p))$ for $M=10$ repetitions of $k=5$ choices, when the guessing rate is purely random, $r=1/k$. In this case, the probability of obtaining 10 correct answers in 10 repetitions of a question, if the model is randomly guessing, is $0.0000001$. 

Conversely, now imagine the model as an oracle which guesses the correct answer at a certain success rate, $SGR$. We can then compute the minimum success needed to always get at least $p$ correct answers in $M$, what we call the \textit{minimum success guessing rate}, $MSGR(p)$. We computed numerically such values, for
the values for $M=10$ repetitions of $k=4$ and $k=5$ choices. For the former, a model has to be guessing at least of a success rate of 0.93 to achieve 6 out of 10 correct answers ($MSGR(6)$), which is equivalent to the requirements of the metric MV proposed in~\cite{pezeshkpour-hruschka-2024-large}.
In our view, a $MSGR(6)= 0.93$ is still insufficient to guarantee that a model is actually consistent in a multiple-choice benchmark. However, requiring that the model is consistent in 10 out of 10 repetitions ($MSGR(10)$) warrants that it can only successfully guess if its success rate is above 0.9999, which for us seem to be an excessive requirement. 

For the experiments in this paper, a model is answer consistent when it is equivalent to an oracle guessing correctly at a $0.99$ rate, or when it has \textit{0.99-answer consistency} or, simply, \textit{0.99-consistency}. As shown by the derivation we have just done, for $M=10$ repetitions of $k=4$ or $k=5$ choices, this is equivalent to answering at least 9 of the 10 repetitions with the same choice. 
This requirement thus covers the two benchmarks used in this paper.

\subsection{Assuring Consistency}\label{sub:assuring-consistency}

We are now in position to propose a method to determine whether a model has 0.99-consistency when answering a 4-~or 5-choice question in a context of 10~repetitions. We say the model is \textit{SURE} of its answers to a question when it shows, experimentally, 0.99-consistency, and \textit{UNSURE} when it is not able to produce evidence of it. This is accomplished by:

\begin{enumerate}
    \item Asking the question to the model 10 times.
    \item If the model answers identically 9 or 10 times, the question is SURE.
    \item If not, the question is UNSURE.
\end{enumerate}

Notice that we do not need to actually call 10~times the model, since in some cases only 3 answers already warranty the model is UNSURE of the question, and, similarly, 9~identical answers are sufficient for SURE. 
However, being SURE does not mean being correct. In the case of benchmarks, where the correct answer is known, we consider that a SURE question is answered correctly, \textit{right}, only if it matches the correct choice for the 9 or 10 times it responds identically, and \textit{wrong} if the converse has happened. In the case of UNSURE questions, we consider here that the model is \textit{right} if the correct answer is among the choices with highest number of answers and \textit{wrong} otherwise.

To characterize differences and trade-offs among models, we propose a representation for \textit{c-consistency} of a model by a pair of numbers separated by the symbol ``\textbar'':
\begin{itemize}
    \item \textbf{RWS}, or ``right when SURE'', is the ratio of right SURE questions to the total number of SURE questions.
    \item \textbf{S/T}, is the percentage of SURE questions in relation to the total number $T$ of questions.
    \item \textbf{c-consistency of a model:} \textbf{RWS \textbar \hspace{0.3mm} S/T}.
\end{itemize}

We understand that repeating a question multiple times is not an efficient way to achieve consistency but, for the purposes of this work, it is a safe method. The proposal of efficient consistency methods is beyond the scope of this paper.



\renewcommand{\arraystretch}{0.9}

\begin{table*}[t!]
\centering
\scriptsize
\begin{tabular}{|l|c|c|c|c|c|c|c|rcl|}
\hline
\textbf{MMLU-Redux - 10 trials} & \ & \tiny{\textbf{SURE}} & \tiny{\textbf{UNSURE}} & \tiny{\textbf{UNSURE}} & \tiny{\textbf{SURE}} & \tiny{\textbf{accuracy}} & \tiny{\textbf{accuracy}} &  &  & \\
\textbf{0.99-consistency} & \tiny{\textbf{temp}} & \tiny{\textbf{\& right}} & \tiny{\textbf{\& right}} & \tiny{\textbf{\& wrong}} & \tiny{\textbf{\& wrong}} & \tiny{\textbf{average}} & \tiny{\textbf{stdev}} & \tiny{\textbf{RWS}} & \textbf{\textbar} & \tiny{\textbf{S/T}} \\ 
\hline
\multicolumn{11}{|c|}{\textbf{SMALL MODELS ($\leq 8B$ parameters)}} \\
\hline
Llama-3-8B & 0.3 & 43\% & 18\% & 29\% & 104\% & \textbf{0.590} & 0.004 & 0.81 &\textbar& \textbf{53\%} \\
Llama-3-8B  & 0.7 & 24\% & 36\% & 39\% & 2\% & 0.520 & 0.005 & 0.93 &\textbar& 25\% \\
Llama-3-8B  & 1.0 & 10\% & 47\% & 43\% & 1\% & 0.428 & 0.006 & \textbf{0.94} &\textbar& 11\% \\
\hline 
Llama-3-8B-instruct & 0.3 & 58\% & 7\% & 14\% & 22\% & \textbf{0.645} & 0.002 & 0.73 &\textbar& \textbf{79\%} \\
Llama-3-8B-instruct  & 0.7 & 49\% & 15\% & 24\% & 12\% & 0.634 & 0.004 & 0.80 &\textbar& 62\% \\
Llama-3-8B-instruct  & 1.0 & 43\% & 21\% & 29\% & 7\% & 0.616 & 0.003 & \textbf{0.86} &\textbar& 50\% \\
\hline
deepseek-llm-7b & 0.3 & 41\% & 8\% & 18\% & 33\% & \textbf{0.485} & 0.002 & 0.55 &\textbar& \textbf{74\%} \\
deepseek-llm-7b  & 0.7 & 32\% & 17\% & 34\% & 17\% & 0.478 & 0.004 & 0.65 &\textbar& 49\% \\
deepseek-llm-7b  & 1.0 & 24\% & 25\% & 42\% & 10\% & 0.465 & 0.003 & \textbf{0.71} &\textbar& 34\% \\

\hline
\multicolumn{11}{|c|}{\textbf{MEDIUM MODELS ($\geq 50B$ and $\leq 80B$ parameters)}} \\
\hline 
llama-3.3-70b & 0.3 & 80\% & 1\% & 1\% & 18\% & \textbf{0.805} & 0.001 & 0.81 &\textbar& \textbf{98\%} \\
llama-3.3-70b & 0.7 & 79\% & 2\% & 2\% & 17\% & 0.805 & 0.001 & 0.82 &\textbar& 96\% \\
llama-3.3-70b & 1.0 & 78\% & 2\% & 4\% & 16\% & 0.805 & 0.001 & \textbf{0.83} &\textbar& 94\% \\ \hline
mixtral-8x7b-instruct & 0.3 & 70\% & 0\% & 1\% & 29\% & \textbf{0.700} & 0.093 & 0.71 &\textbar& \textbf{99\%} \\
mixtral-8x7b-instruct & 0.7 & 70\% & 0\% & 1\% & 29\% & 0.687 & 0.001 & 0.71 &\textbar& 98\% \\
mixtral-8x7b-instruct & 1.0 & 69\% & 1\% & 2\% & 29\% & 0.679 & 0.105 & \textbf{0.71} &\textbar& 98\% \\ \hline
qwen2-5-72b-instruct & 0.3 & 81\% & 2\% & 2\% & 15\% & \textbf{0.826} & 0.001 & 0.84 &\textbar& 96\% \\
qwen2-5-72b-instruct & 0.7 & 79\% & 4\% & 5\% & 12\% & 0.825 & 0.001 & 0.87 &\textbar& 91\% \\
qwen2-5-72b-instruct & 1.0 & 77\% & 6\% & 7\% & 10\% & 0.823 & 0.001 & \textbf{0.88} &\textbar& 87\% \\

\hline
\multicolumn{11}{|c|}{\textbf{SMALL GRANITE MODELS ( $\leq 8B$ parameters)}} \\
\hline 
3.2-8b-instruct & 	0.3 &	56\% &	8\% &	14\% &	22\% & \textbf{0.636}	& 0.002	& 0.72	&\textbar&	\textbf{78\%} \\
3.2-8b-instruct &	0.7	& 48\% &	16\% &	24\% &	12\% & 0.626 & 0.004	& 0.80	&\textbar&	60\%  \\
3.2-8b-instruct &	1.0	& 40\% &	24\% &	28\% &	8\% & 0.611	& 0.003	& \textbf{0.84}	&\textbar&	48\%   \\
\hline
3.1-8b-instruct &	0.3	& 56\% &	8\% &	14\% &	21\% 	& \textbf{0.640}	& 0.002	& 0.71	&\textbar&	\textbf{78\%}  \\
3.1-8b-instruct &	0.7	& 46\% &	18\% &	25\% &	11\% 	& 0.624	& 0.002	& 0.81&\textbar&	57\%  \\
3.1-8b-instruct &	1.0	& 37\% &	26\% &	30\% &	7\% 	& 0.592	& 0.006	& \textbf{0.84}	&\textbar&	44\%  \\
\hline
3.1-8b-base &	0.3	& 44\% &	14\% &	24\% &	18\% & \textbf{0.575}	& 0.002 & 0.71	&\textbar&	\textbf{63\%}  \\
3.1-8b-base &	0.7	& 24\% &	34\% &	38\% &	4\% & 0.535 & 0.005	& 0.84 &\textbar&	28\%  \\
3.1-8b-base &	1.0	& 10\% &	48\% &	41\% &	11\% & 0.462 & 0.007 & \textbf{0.90} &\textbar&	11\%  \\
\hline
3.0-8b-base &	0.3	& 42\% &	19\% &	28\% &	11\% & \textbf{0.591}	& 0.003 & 0.79	&\textbar&	\textbf{52\%}  \\
3.0-8b-base &	0.7	& 21\% &	39\% &	37\% &	2\% & 0.537	& 0.005	& 0.90	&\textbar&	24\%  \\
3.0-8b-base &	1.0	& 10\% &	49\% &	40\% &	1\% & 0.457	& 0.004	& \textbf{0.90}	&\textbar&	11\%  \\
\hline
3.1-2b-instruct &	0.3	& 56\% &	8\% &	14\% &	22\% & \textbf{0.636}	& 0.002	& 0.72	&\textbar&	\textbf{78\%}  \\
3.1-2b-instruct &	0.7	& 48\% &	16\% &	24\% &	12\% & 0.626	& 0.004	& 0.80	&\textbar&	60\%  \\
3.1-2b-instruct &	1.0	& 40\% &	24\% &	28\% &	8\% & 0.611 & 0.003 & \textbf{0.84}	&\textbar&	48\%  \\
\hline
3.1-2b-base &	0.3	& 21\% &	28\% &	44\% &	7\% & \textbf{0.465}	& 0.005 & 0.76	&\textbar&	\textbf{24\%}  \\
3.1-2b-base &	0.7	& 7\% &	41\% &	52\% &	1\% & 0.394	& 0.006 & 0.91 &\textbar&	7\%  \\
3.1-2b-base &	1.0	& 2\% &	42\% &	55\% &	0\% & 0.331	& 0.004	& \textbf{0.91} &\textbar&	3\%  \\

\hline
\end{tabular}
\caption{Results of 0.99-consistency for the MMLU-Redux benchmark with 10 repetitions for 3~small models  (2B-8B) , 3~medium models  (50B-80B) , and for 6~small models of the granite family.}
\label{tab:mmlu-results}
\end{table*}

\section{Experiments with MMLM-Redux}\label{mmlu-results}

We first explored answer consistency by running 10 repetitions of the \textbf{MMLU-Redux} general knowledge
benchmark~\cite{gema2025mmlu}, considering three general-purpose small models in the range of 8~billion parameters: \textit{LLama3 8B} (Llama-3-8B)~\cite{grattafiori2024llama3herdmodels}, \textit{LLama3 8B Instruct} (Llama-3-8B-instruct)~\cite{grattafiori2024llama3herdmodels}, and \textit{DeepSeek v3 7B}~\cite{deepseekai2024deepseekv3technicalreport}. 

We also explored three medium-size models, \textit{Llama3 v3 70B} (llama-3.3-70b)~\cite{grattafiori2024llama3herdmodels}, \textit{Mistral 8x7B Instruct} (mixtral-8x7b-instruct)~\cite{jiang2024mixtralexperts}, and \textit{Qwen 2.5 72B Instruct} (qwen2-5-72b-instruct)~\cite{qwen2025qwen25technicalreport}, all in the range of 50~to 70~billion parameters. The labeling of the models as \textit{small} and \textit{medium} follows practices of the LLM research community.

The experiment consisted on performing 30~repetitions of the evaluation with the benchmark, in three sets of 10~identical repetitions, with \textit{inference temperatures} of $0.3$, $0.7$, and $1.0$. MMLU-Redux is a 4-choice benchmark, so we computed the correct answer by prompting the model to answer the question as the letter of the correct choice, using the prompt described next, with top-K sampling decoding. 

\subsection{Prompt Development and Runtime Inference Details}\label{sec:prompt}

Both benchmarks used for this work, namely MMLU-Redux and MedQA, are multiple-choice benchmarks, which compose a subset of evaluation types where we can very objectively compute the correct answer by prompting the model to answer the question as the letter of the correct choice. For each question, we formatted the prompt below by replacing the $QUESTION$ and $CHOICES$ variables with the question and corresponding alternatives. This combination of prompt plus a set of inference parameters was chosen after initial trial-and-error tests. 

\begin{small}
\begin{verbatim}
Answer the following multiple choice 
questions. The first line of your 
response should be: 'LETTER' (without 
quotes), followed by a 
step-by-step explanation.

Question: $QUESTION$
Choices: $CHOICES$
Answer:
\end{verbatim}
\end{small}

\begin{figure}[th!]   
     \centering
      \includegraphics[width=1.0\linewidth]{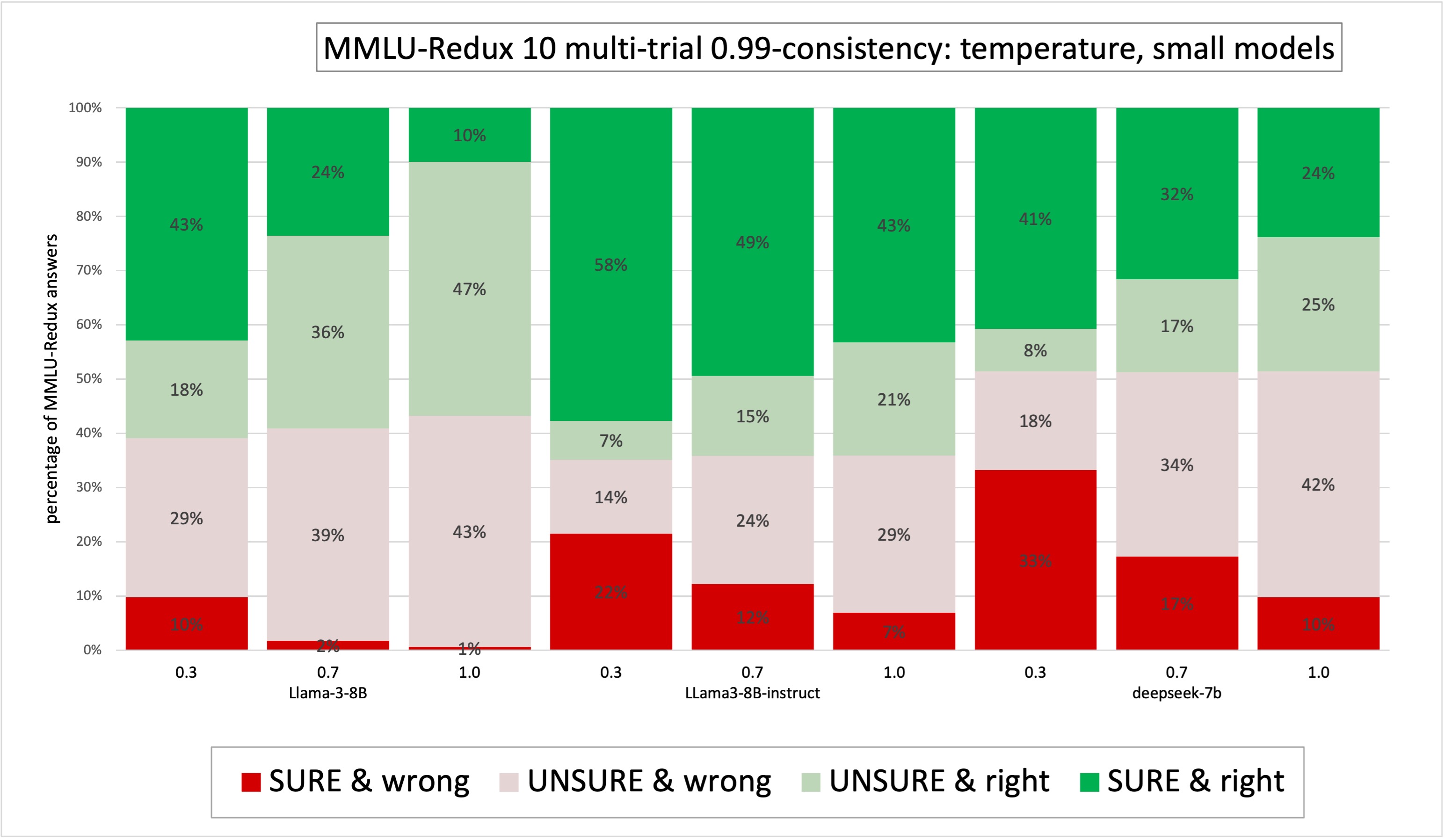}
     \caption{Bar graph showing the proportions of SURE/UNSURE and right/wrong of the 3~small models for the MMLU-Redux benchmark, 0.99-consistency, for 3 temperatures.} 
     \label{fig:app-mmlu-temperature-small}
\end{figure}

\begin{figure}[th!]   
     \centering
      \includegraphics[width=1.0\linewidth]{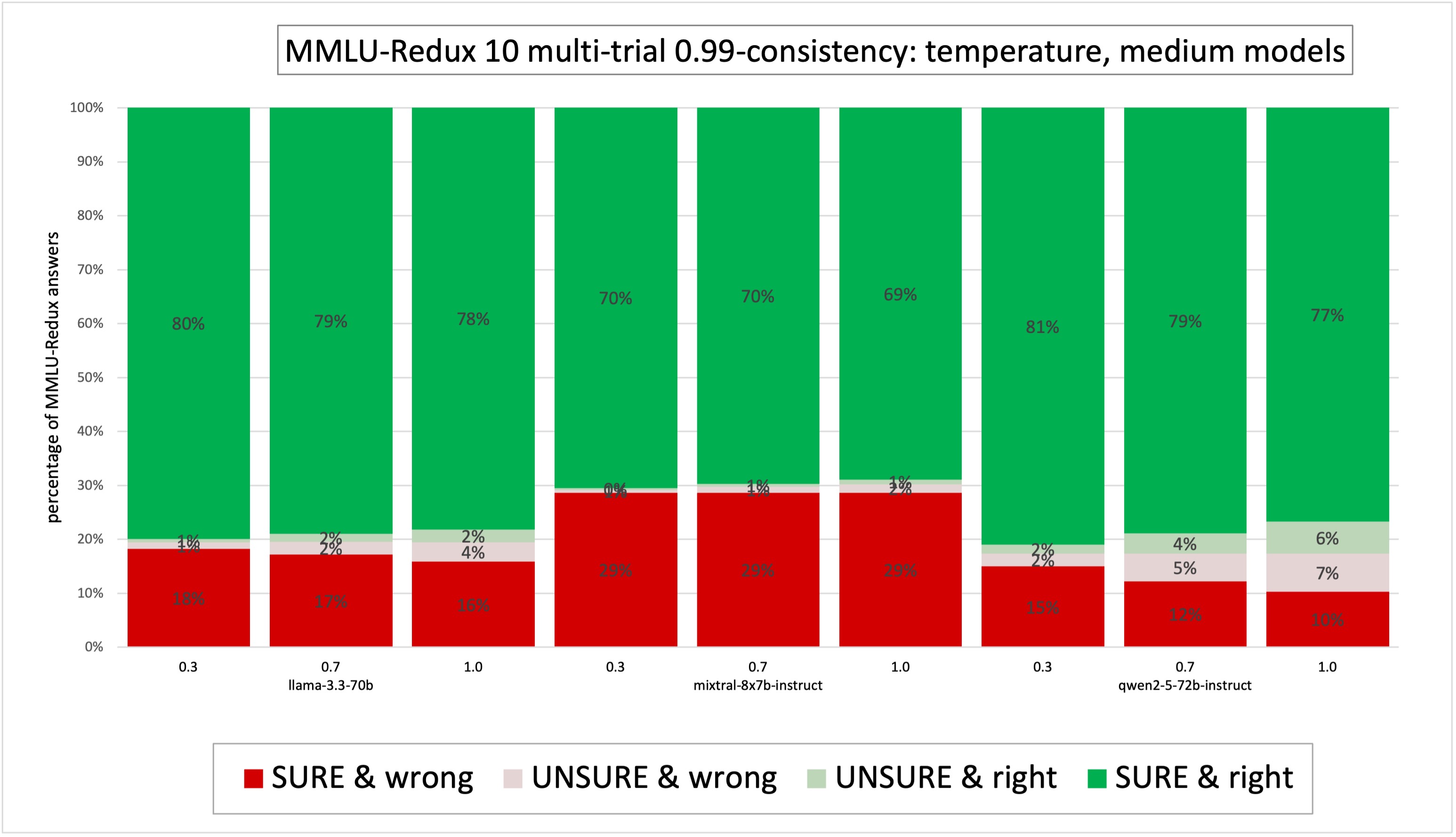}
     \caption{Bar graph showing the proportions of SURE/UNSURE and right/wrong of the 3~medium models for the MMLU-Redux benchmark, 0.99-consistency, for 3 temperatures.} 
     \label{fig:app-mmlu-temperature-medium}
\end{figure}

\begin{figure*}[th!]   
     \centering
      \includegraphics[width=0.8\linewidth]{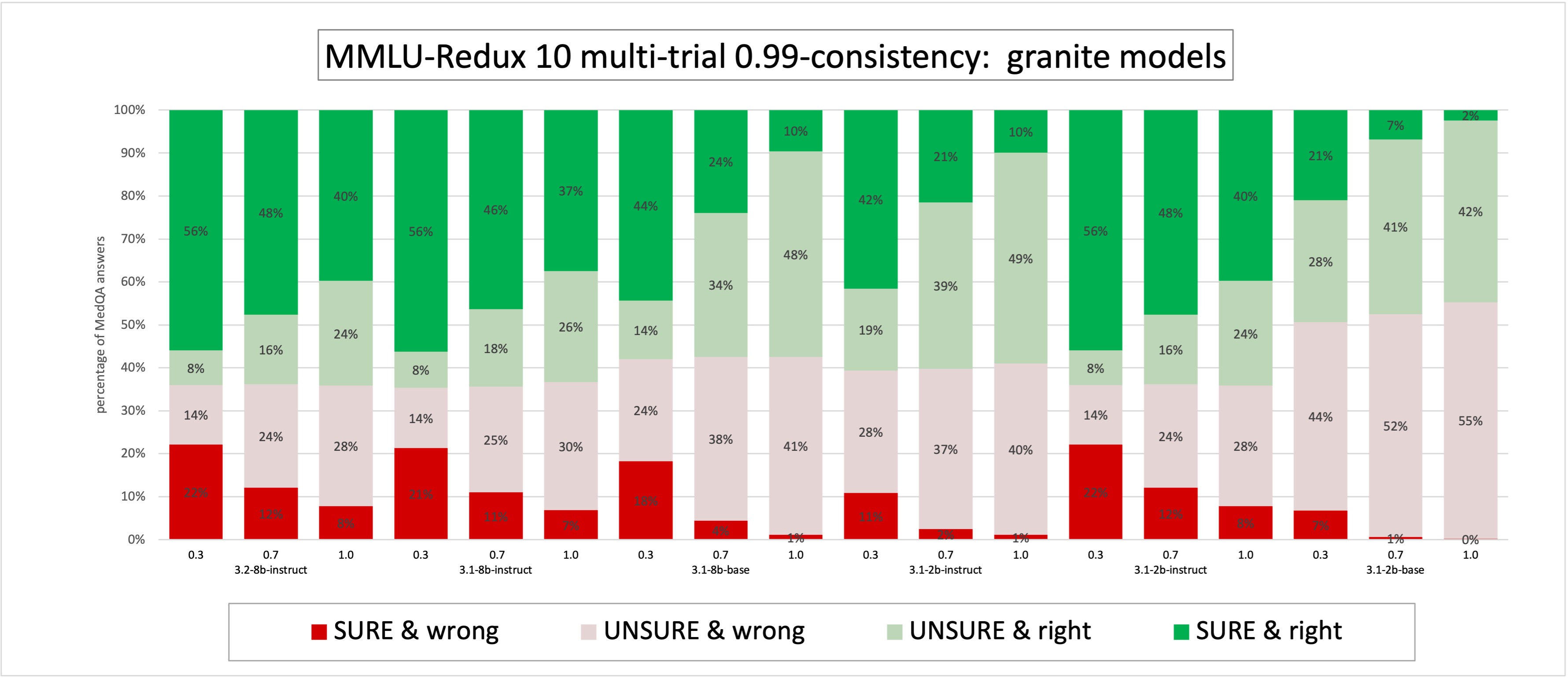}
     \caption{Bar graph showing the proportions of SURE/UNSURE and right/wrong of the 3~granite models for the MMLU-Redux benchmark, 0.99-consistency, for 3 temperatures.} 
     \label{fig:app-mmlu-temperature-granite}
\end{figure*}

Given an answer, we parsed the inferred text to extract the choice generated by the model. To consider a question as correct or not, we then compared the generated choice to the correct choice. As stated previously, for this case study we used top-K sampling decoding~\cite{song2024good}.

The models inference was conducted using HuggingFace \verb|transformers| Python library, running on an internal infrastructure using a single A100 GPU with 80GB of memory. The only exception was for the Llama3 v3 70B model, for which we accessed this model hosted internally in a more scalable infrastructure with multiple GPUs through an available API. Technical details about the models are provided in section~\ref{app:references-models} of the Appendix.

In terms of parameters, we employed the default top-K search inference parameters provided by \verb|transformers| and adjusted the temperature parameter for each experiment, using 0.3, 0.7, and 1.0 as temperature values. We ran most experiments with a default \verb|max_new_tokens| value of 3. Although we have  experimented with a value of 5 and 256 tokens for MMLU-Redux and, since we were only targeting the choice provided by the model, we reduced \verb|max_new_tokens| to 3 to avoid unnecessary computations.
Considering the internal infrastructure with A100 GPUs, the average duration for generating answers were between 1 to 4 days, depending on the benchmark and model parameter combination, such as different choices of \verb|max_new_tokens|, number of repetitions, and temperature values. In total, these experiments to about 3 to 5 days to run.




\subsection{Small vs. Medium Models}

Results for the 3~small and 3~medium models are shown on the top part of table~\ref{tab:mmlu-results}. For each evaluation we computed the ratio of questions for which the model provided the correct alternative, yielding 10~results of accuracy. The average of those 10~accuracy results is reported as the \textit{accuracy average} together with the standard deviation, \textit{accuracy stdev}.

We then determined, examining the 10 answers for each question in an evaluation, whether the model answered the question in a SURE or UNSURE way and whether the set of 10~answers was right or wrong as described before. We followed by determining the ratio of correct questions to the number of SURE questions, RWS, and the percentage of SURE questions, S/T, as described previously.

The results of the experiment with small and medium models are depicted on the top of table~\ref{tab:mmlu-results}. Figure~\ref{fig:app-mmlu-temperature-small} depicts the bar graphs showing the proportions of SURE/UNSURE and right/wrong of the 3 small models for the MMLU-redux benchmark, at 0.99-consistency, for 3 temperatures and, similarly, figure~\ref{fig:app-mmlu-temperature-medium} depicts the results for the medium models. 


For small models, the best temperature was $t=0.3$, in which the percentage of SURE questions (S/T) was in the 53\% to 79\% range and the accuracy on consistent answers (RWS) was  better than the average accuracy. Notice also a pattern where the best S/T scores were at the low temperatures while the best RWSs happened at high temperatures.

 The medium models had very high levels of percentage of SURE questions (S/T), from 96\% to 99\% at their best temperatures, suggesting that bigger models may have less issues with consistency. Possibly because of that, their accuracy average and RWS scores were very similar, and we saw again the best RWS scores happening at high temperatures. These differences are clearly seen when comparing the bar graphs of figures~\ref{fig:app-mmlu-temperature-small} and~\ref{fig:app-mmlu-temperature-medium}. However, bigger models have a higher chance of benchmark contamination, since they are trained with much larger amounts of data. We will further explore possible contamination issues in large models in future works.

\subsection{Exploring Multiple Versions of a Model}\label{mmlu-granite-results}

We also performed an identical study using 6~different models of the \textit{granite} family~\cite{granite2024granite}: \textit{3.2 8B instruct} (3.2-8b-instruct), \textit{3.1 8B Instruct} (3.1-8b-instruct), \textit{3.1 8B Base} (3.1-8b-base), \textit{3.0 8B Base} (3.0-8b-base \textit{3.1 2B Instruct} (3.1-2b-instruct), and \textit{3.1 2B Base} (3.1-2b-base), comprising models with between 2~and 8~billion parameters. 
We used the granite models because they were trained with carefully curated data and with a strong attention to avoid incorporating benchmarks~\cite{granite2024granite}.

Results are shown on table~\ref{tab:mmlu-results} and the corresponding bar graph is shown in figure~\ref{fig:app-mmlu-temperature-granite}. We observed very similar results: the percentage of SURE questions at the best temperature for each model going from 52\% to 75\% (we ignored the results of the 3.1-2b-base as outliers); the accuracy among SURE questions (RWS) a little higher then the average accuracy; and the best performance for S/T at $t=0.3$. Similarly, the best RWS scores were at high temperatures:  a trade-off between being SURE often and being right when SURE.




\begin{figure}[th!]   
     \centering
      \includegraphics[width=0.9\linewidth]{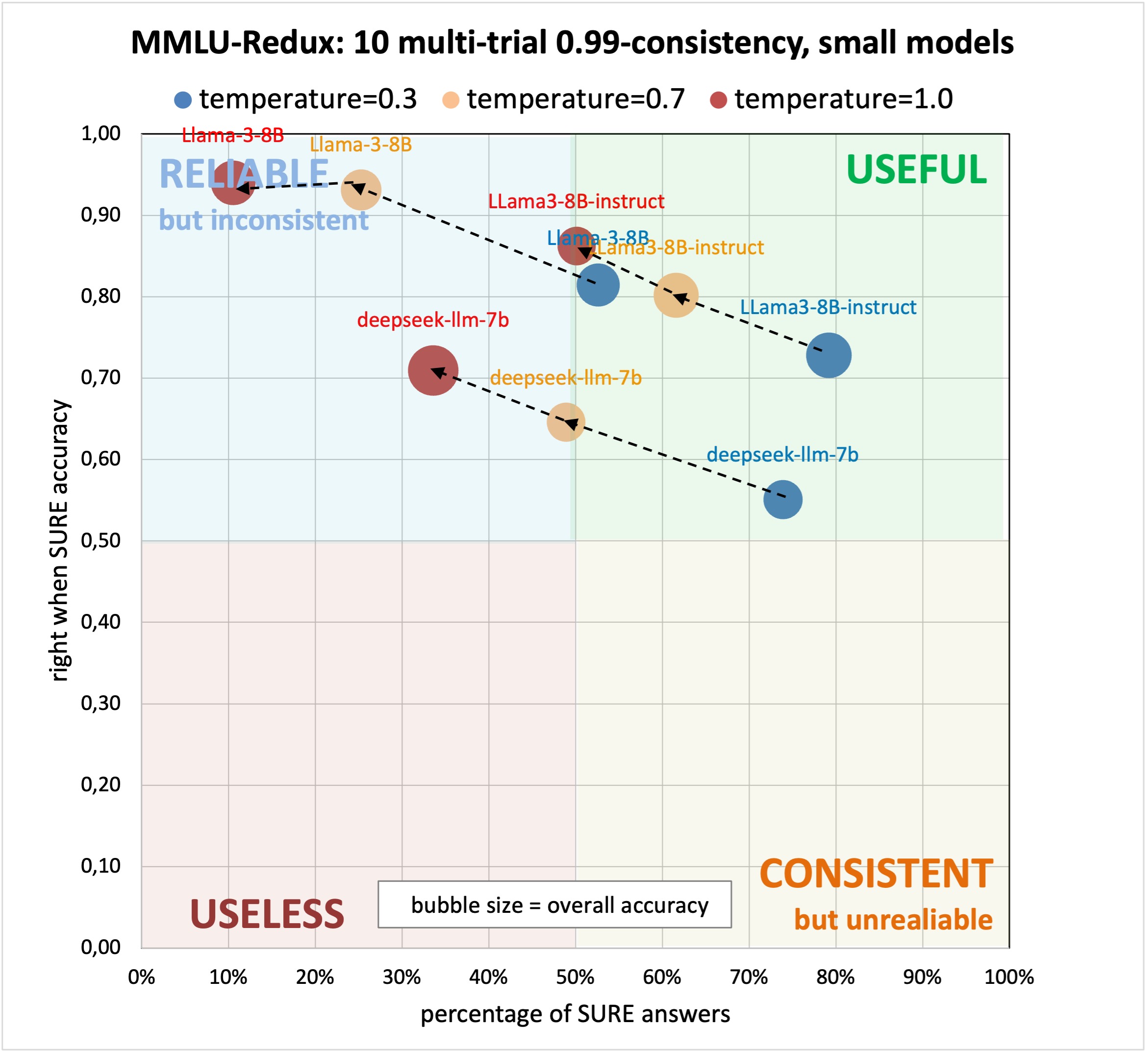} \\
      \vspace{2mm}
      \includegraphics[width=0.9\linewidth]{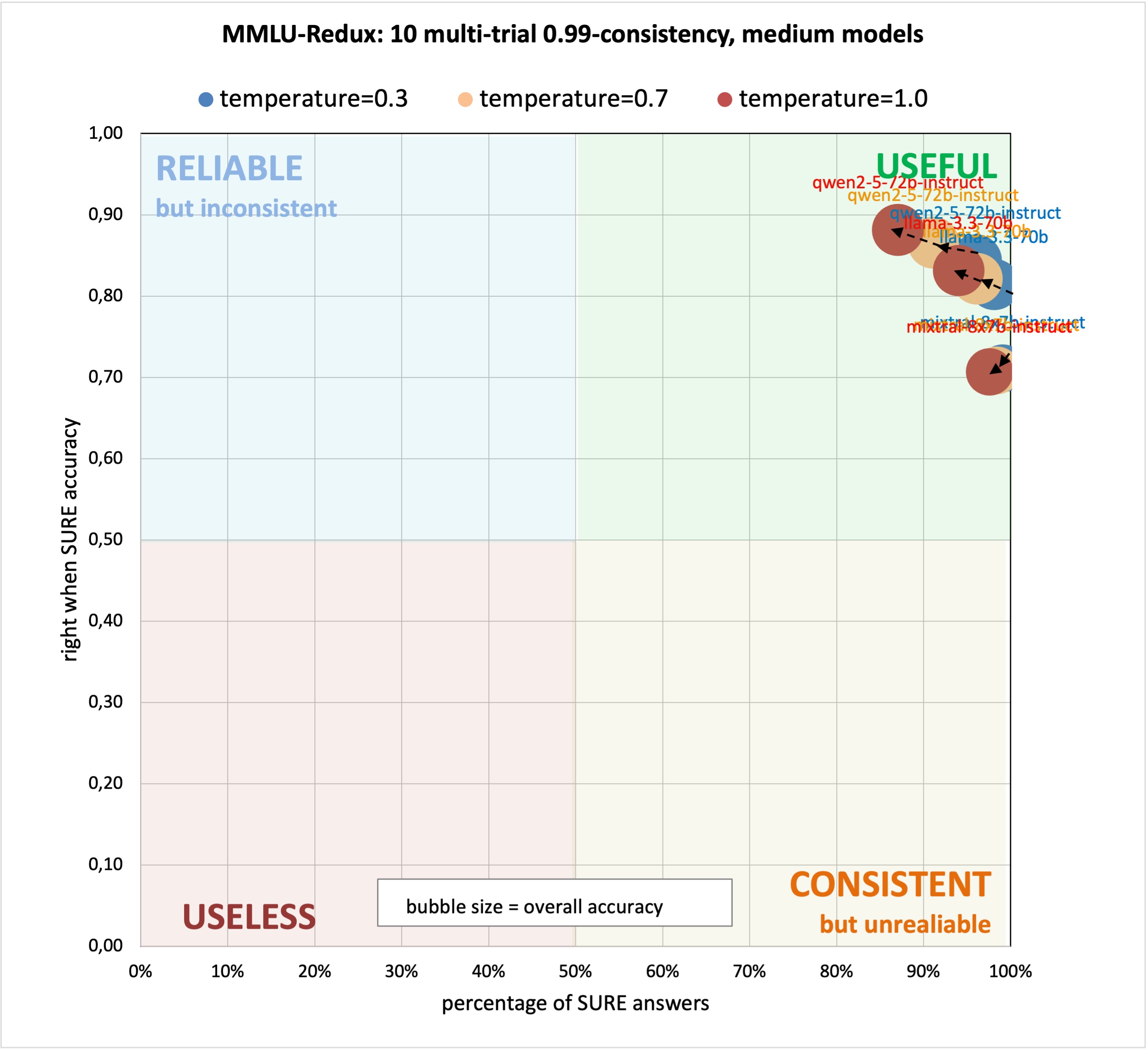} \\
      \vspace{2mm}
      \includegraphics[width=0.9\linewidth]{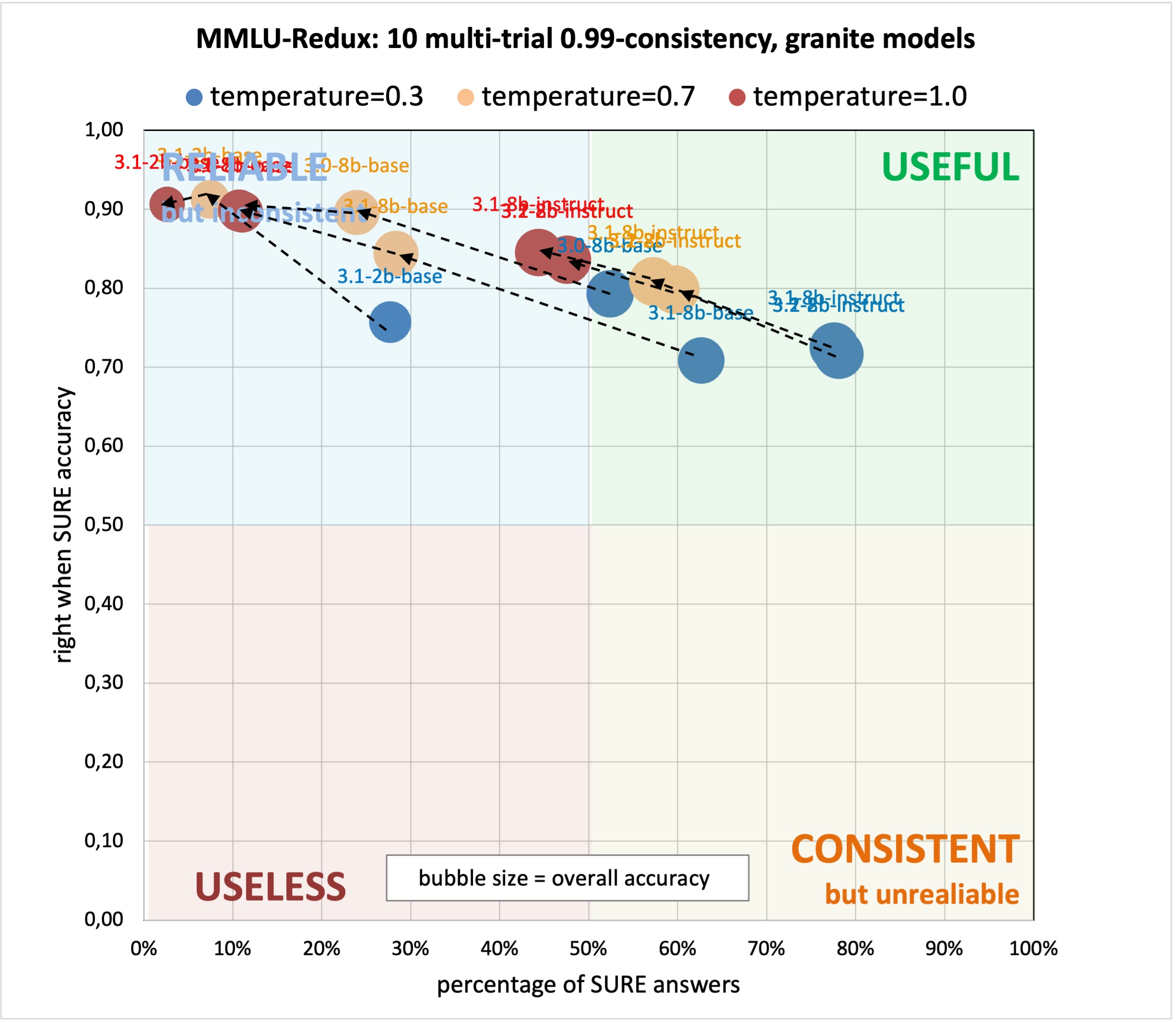}
     \caption{Consistency plots for the MMLU-Redux benchmark for small (top), medium (center), and granite (bottom) models.} 
     \label{fig:app-mmlu-consistency-plots}
\end{figure}

\subsection{Visualization with Consistency Plots}\label{sec:consistency-plot}

To help visualize and compare different levels of answer consistency, we created a graph we call the \textit{consistency plot}. An example is shown on top of figure~\ref{fig:app-mmlu-consistency-plots}, depicting the 0.99-consistency numbers of the of the 3 small models according to the results shown on table~\ref{tab:mmlu-results}.
The consistency plot uses the X axis to represent S/T , that is, the percentage of SURE answers in relation to the total number $T$ of questions; 
and the Y axis to represent RWS, the ratio of right answers to the total number of SURE questions as defined in section~\ref{sub:assuring-consistency}. 
The overall accuracy of the model, for all $T$ questions in all repetitions, is associated to the area of the circle.

In the consistency plot shown on the top of figure~\ref{fig:app-mmlu-consistency-plots}, we see that all three models follow a distinct path, right-to-left, bottom-to-top as the temperature increases. To simplify the reading of these paths, we named the four quadrants with their most salient characteristics relatively to consistency:
\begin{itemize}
    \item \textbf{Top Right: USEFUL}, of models which have both a higher level of consistent answers and which are often right when they are consistent;
    \item \textbf{Top Left: RELIABLE but inconsistent}, of models which less than half of time produce consistent answers but are often correct in those answers;
    \item \textbf{Bottom Right: CONSISTENT but unreliable}, of models which often answer consistently, but are often incorrect on those answers;
    \item \textbf{Bottom Left: USELESS}, of models which are both often inconsistent and incorrect.
\end{itemize}

The middle and bottom graphs of figure~\ref{fig:app-mmlu-consistency-plots} show the consistency plots of medium and granite models, respectively. In the case of the medium models, all the models are very close to each other on the top right of the graph, almost independently of the inference temperature. This helps visualizing how high levels of answer consistency and reliability are depicted in consistency plots. In the case of the granite models (bottom of figure~\ref{fig:app-mmlu-consistency-plots}), the same patterns of the small models (top of of figure~\ref{fig:app-mmlu-consistency-plots}) are seen, namely, paths which go right-to-left, bottom-to-top as the temperature increases, although with higher accuracy among SURE answers, RWS.

\subsection{Aggregated Results over All Models }

\begin{figure}[th!]   
    \centering
     \includegraphics[width=0.95\linewidth]{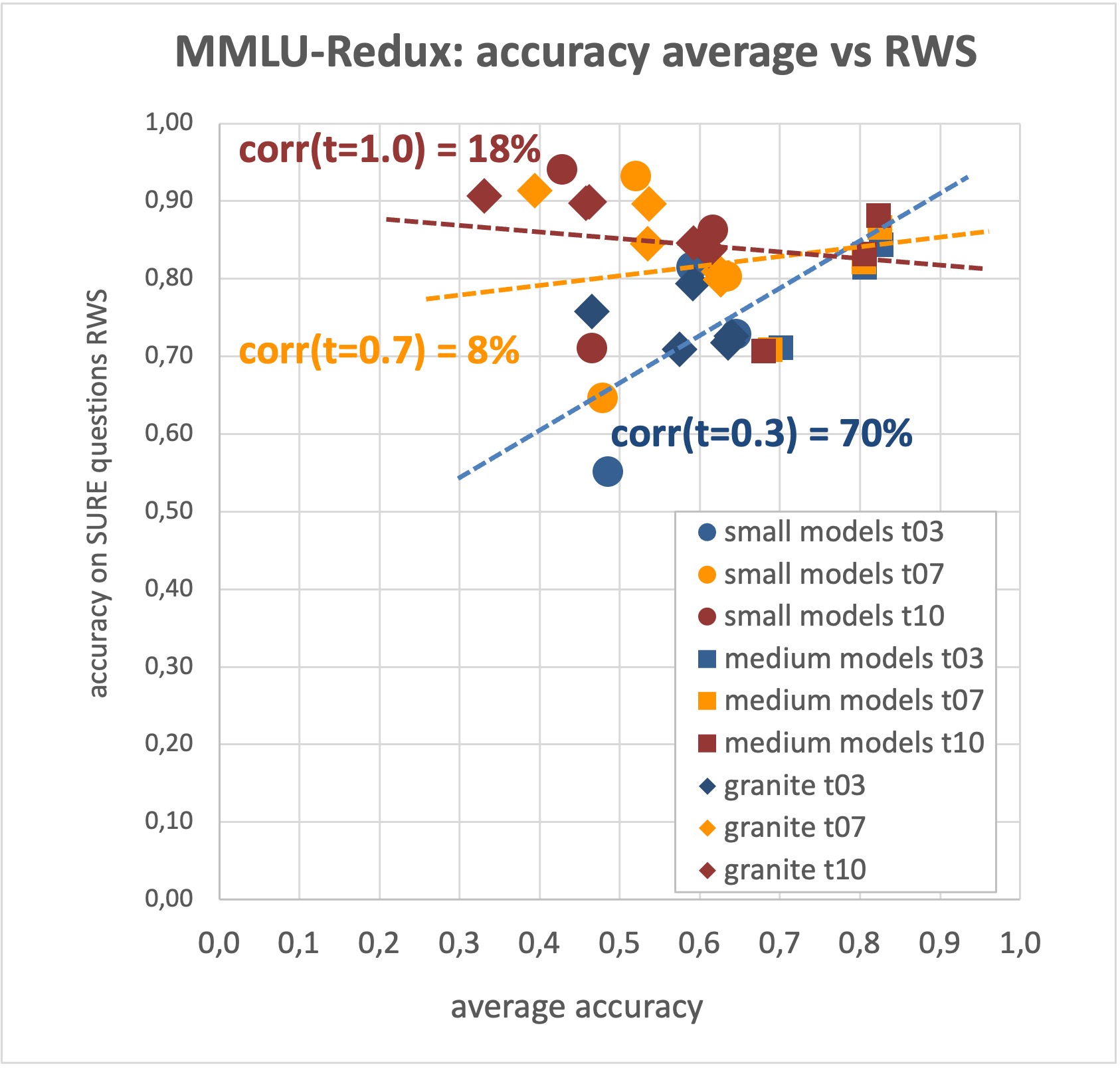} \\
     \vspace{1mm}
     \includegraphics[width=0.95\linewidth]{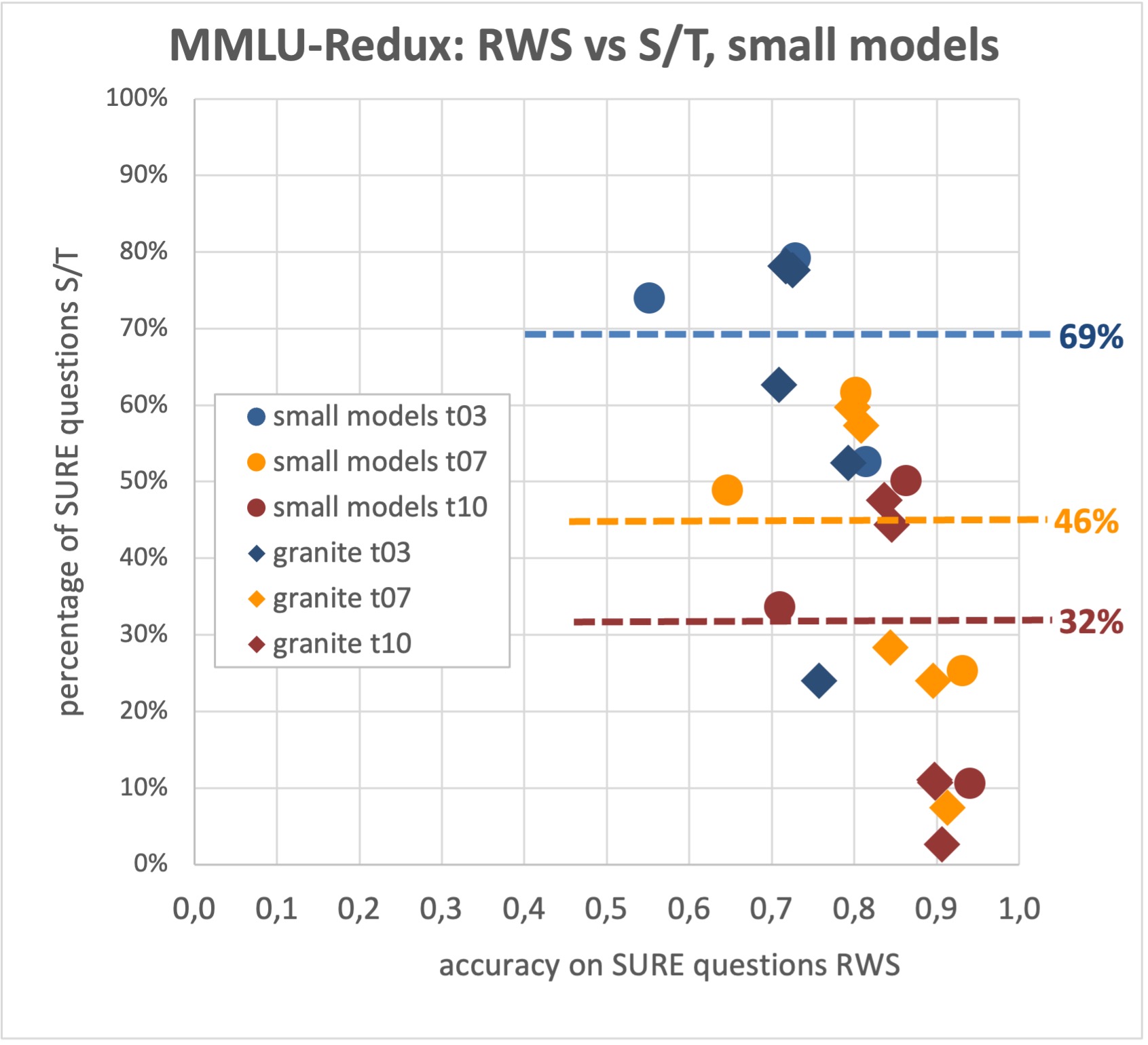}
    \caption{Scatter plot graphs showing the correlation between average accuracy and RWS for small and medium models (top) and between RWS and S/T  for small models (bottom) tested with MMLU-Redux benchmark.} 
    \label{fig:mmlu-aggregated}

\end{figure}

To provide an overall understanding of the impact of assuring consistency, we first considered the RWS results of all models at different temperatures and computed the correlation with the average accuracy. We obtained correlation values of 70\%, 8\%, and 18\% for the results with temperatures 0.3, 0.7, and 1.0, respectively. In the scatter plot depicted on the top of figure~\ref{fig:mmlu-aggregated}, we can see that the blue markers corresponding to the evaluations with $t=0.3$ are reasonably aligned. Using basic regression, we obtained a coefficient of $0.563$ and an intercept value of $0.374$, with $R^2=0.488$. 

Similarly, we considered the percentage of SURE questions (S/T) for each model at the $t=0.3$ for all 9~small models and obtained an average across models of 69\% with an standard deviation of 11\%, therefore exhibiting a tendency to be in the range between 58\% and 80\%. We did not consider the medium models because, as observed before, they exhibit very strong consistency, in the range of the upper 90\%. For $t=0.7$, we obtained an average of 46\% $\pm$ 17\% and for $t=1.0$,  32\% $\pm$ 18\% (see bottom of figure~\ref{fig:mmlu-aggregated}).

This seems to indicate that small models, even at the low temperature 0.3, tend to exhibit high levels of inconsistency: in none of our experiments with MMLU-Redux, any of them showed more than 80\%.
The scatter plot on the bottom of figure~\ref{fig:mmlu-aggregated} shows RWS values also stayed in a small range, from 0.55 to 0.94.

\section{Experiments with MedQA}\label{results-medqa}

\begin{table*}[t!]
\centering
\scriptsize
\begin{tabular}{|l|c|c|c|c|c|c|c|rcl|}
\hline

\textbf{MedQA - 10 trials} & \ & \tiny{\textbf{SURE}} & \tiny{\textbf{UNSURE}} & \tiny{\textbf{UNSURE}} & \tiny{\textbf{SURE}} & \tiny{\textbf{accuracy}} & \tiny{\textbf{accuracy}} &  &  & \\
\textbf{0.99-consistency} & \tiny{\textbf{temp}} & \tiny{\textbf{\& right}} & \tiny{\textbf{\& right}} & \tiny{\textbf{\& wrong}} & \tiny{\textbf{\& wrong}} & \tiny{\textbf{average}} & \tiny{\textbf{stdev}} & \tiny{\textbf{RWS}} & \textbf{\textbar} & \tiny{\textbf{S/T}} \\ 
\hline
\multicolumn{11}{|c|}{\textbf{SMALL FINETUNED MODELS}} \\
\hline
medllama3-v20 & 0.3 & 71\% & 8\% & 6\% & 24\% & \textbf{0.736} & 0.002 & 0.75 &\textbar& \textbf{96\%} \\
medllama3-v20  & 0.7 & 67\% & 6\% & 5\% & 21\% & 0.733 & 0.004 & \textbf{0.76} &\textbar& 89\% \\
medllama3-v20  & 1.0 & 67\% & 7\% & 5\% & 21\% & 0.730 & 0.004 & \textbf{0.76} &\textbar& 88\% \\
\hline
Bio-Medical-Llama-3-8B  & 0.3 & 66\% & 7\% & 5\% & 23\% & \textbf{0.717} & 0.003 & 0.74 &\textbar& \textbf{88\%} \\
Bio-Medical-Llama-3-8B  & 0.7 & 47\% & 25\% & 15\% & 13\% & 0.701 & 0.007 & \textbf{0.79} &\textbar& 60\% \\
Bio-Medical-Llama-3-8B  & 1.0 & 48\% & 23\% & 16\% & 13\% & 0.691 & 0.007 & \textbf{0.79}  &\textbar& 61\% \\
\hline
BioMistral-7B & 0.3 & 25\% & 12\% & 14\% & 49\% & \textbf{0.371} & 0.004 & \textbf{0.34} &\textbar& \textbf{75\%} \\
BioMistral-7B & 0.7 & 7\% & 26\% & 40\% & 27\% & 0.329 & 0.011 & 0.20 &\textbar& 34\% \\
BioMistral-7B  & 1.0 & 7\% & 25\% & 40\% & 28\% & 0.327 & 0.011 & 0.02 &\textbar& 35\% \\
\hline
medalpaca-7b  & 0.3 & 14\% & 18\% & 32\% & 35\% & \textbf{0.342} & 0.012 & \textbf{0.29} &\textbar& \textbf{49\%} \\
medalpaca-7b & 0.7 & 2\% & 21\% & 60\% & 17\% & 0.289 & 0.015 & 0.11 &\textbar& 19\% \\
medalpaca-7b & 1.0 & 2\% & 20\% & 64\% & 14\% & 0.290 & 0.008 & 0.14 &\textbar& 17\% \\
\hline
\multicolumn{11}{|c|}{\textbf{SMALL BASE MODELS}} \\
\hline
Llama3-8B & 0.3 & 38\% & 14\% & 14\% & 34\% & \textbf{0.512} & 0.008 & \textbf{0.53} &\textbar& \textbf{72\%}\\
Llama3-8B  & 0.7 & 19\% & 30\% & 31\% & 20\% & 0.472 & 0.006 & 0.49 &\textbar& 39\% \\
Llama3-8B  & 1.0 & 10\% & 33\% & 43\% & 14\% & 0.242 & 0.007 & 0.41 &\textbar& 25\% \\
\hline
Llama3-8B-instruct  & 0.3 & 50\% & 8\% & 6\% & 37\% & 0.472 & 0.006 & 0.49 &\textbar& \textbf{87\%} \\
Llama3-8B-instruct  & 0.7 & 42\% & 15\% & 15\% & 29\% & \textbf{0.570} & 0.005 & 0.58 &\textbar& 70\% \\
Llama3-8B-instruct  & 1.0 & 33\% & 23\% & 22\% & 22\% & 0.541 & 0.004 & \textbf{0.60} &\textbar& 55\% \\
\hline
Mistral-7B-Instruct  & 0.3 & 23\% & 10\% & 11\% & 54\% & \textbf{0.321} & 0.008 & \textbf{0.29} &\textbar& \textbf{79\%} \\
Mistral-7B-Instruct  & 0.7 & 14\% & 16\% & 26\% & 44\% & 0.310 & 0.009 & 0.24 &\textbar& 58\% \\
Mistral-7B-Instruct  & 1.0 & 8\% & 21\% & 36\% & 35\% & 0.300 & 0.010 & 0.19 &\textbar& 43\% \\
\hline
llama1-7b  & 0.3 & 0\% & 9\% & 70\% & 20\% & \textbf{0.201} & 0.009 & \textbf{0.00} &\textbar& \textbf{20\%} \\
llama1-7b  & 0.7 & 0\% & 6\% & 78\% & 16\% & 0.187 & 0.014 & 0.00 &\textbar& 16\% \\
llama1-7b  & 1.0 & 0\% & 5\% & 77\% & 18\% & 0.165 & 0.012 & 0.00 &\textbar& 18\% \\
\hline
\multicolumn{11}{|c|}{\textbf{SMALL GRANITE MODELS}} \\
\hline
3.2-8b-instruct & 	0.3 &	30\% &	13\% &	13\% &	44\% & \textbf{0.422}	& 0.005	& \textbf{0.41}	&\textbar&	\textbf{74\%} \\
3.2-8b-instruct &	0.7	& 19\% &	21\% &	28\% &	32\% & 0.393	& 0.008	& 0.38	&\textbar&	51\%  \\
3.2-8b-instruct &	1.0	& 11\% &	27\% &	38\% &	25\% & 0.376	& 0.007	& 0.32	&\textbar&	36\%   \\
\hline
3.1-8b-instruct &	0.3	& 32\% &	12\% &	12\% &	44\% 	& \textbf{0.433}	& 0.006	& \textbf{0.42}	&\textbar&	\textbf{75\%}  \\
3.1-8b-instruct &	0.7	& 19\% &	22\% &	28\% &	30\% 	& 0.405	& 0.007	& 0.39	&\textbar&	49\%  \\
3.1-8b-instruct &	1.0	& 12\% &	26\% &	36\% &	26\% 	& 0.368	& 0.011	& 0.32	&\textbar&	39\%  \\
\hline

3.1-8b-base &	0.3	& 11\% &	23\% &	37\% &	29\% & \textbf{0.349}	& 0.008	& \textbf{0.27}	&\textbar&	\textbf{40\%}  \\
3.1-8b-base &	0.7	& 2\% &	21\% &	60\% &	16\% & 0.283	& 0.007	& 0.13	&\textbar&	18\%  \\
3.1-8b-base &	1.0	& 0\% &	17\% &	67\% &	15\% & 0.234	& 0.010	& 0.00	&\textbar&	16\%  \\
\hline
3.0-8b-base &	0.3	& 17\% &	23\% &	49\% &	11\% & \textbf{0.369}	& 0.008	& 0.60	&\textbar&	\textbf{29\%}  \\
3.0-8b-base &	0.7	& 4\% &	33\% &	61\% &	1\% & 0.293	& 0.010	& 0.75&\textbar&	5\%  \\
3.0-8b-base &	1.0	& 1\% &	36\% &	63\% &	0\% & 0.244	& 0.010 & \textbf{0.79} &\textbar&	1\%  \\
\hline
3.1-2b-instruct &	0.3	& 22\% &	14\% &	11\% &	53\% & \textbf{0.296}	& 0.011	& \textbf{0.29}	&\textbar&	\textbf{75\%}  \\
3.1-2b-instruct &	0.7	& 11\% &	21\% &	27\% &	40\% & 0.263	& 0.010	& 0.22	&\textbar&	52\%  \\
3.1-2b-instruct &	1.0	& 6\% &	22\% &	38\% &	34\% & 0.244	& 0.009	& 0.15	&\textbar&	40\%  \\
\hline
3.1-2b-base &	0.3	& 9\% &	19\% &	33\% &	39\% & \textbf{0.271}	& 0.010	& \textbf{0.19}	&\textbar&	\textbf{49\%}  \\
3.1-2b-base &	0.7	& 1\% &	19\% &	59\% &	21\% & 0.239	& 0.011	& 0.04	&\textbar&	21\%  \\
3.1-2b-base &	1.0	& 0\% &	14\% &	68\% &	18\% & 0.217	& 0.010	& 0.00	&\textbar&	19\%  \\
\hline

\end{tabular}
\caption{Results of 0.99-consistency of models on the MedQA benchmark with 10 repetitions for 4~finetuned models, 4~base models, and for 6~models of the granite family.}
\label{tab:medqa-results}
\end{table*}

To explore whether those results were particular to the MMLU-Redux benchmark, we ran similar studies with the health-related \textit{MedQA} benchmark~\cite{jin2021disease} on 4~medical models and their respective base models, again using 3~different temperatures. Instead of exploring the differences between small and medium models, as we did with the MMLU-Redux benchmark, we explored the differences in answer consistency among \textit{finetuned} vs. \textit{base} models. The objective was to see whether and how the finetuning process may affect consistency. We also performed evaluations with the same 6~granite models, taking in account that MedQA is beyond their areas of expertise.

Here, we considered the test set of MedQA with 1,273 5-choice questions extracted from the USMLE exam. We chose 4~LLMs finetuned on medical data which had reported scores on this benchmark: \textit{MedLlama3 7B} (medllama3-v20)~\cite{ProbeMedicalYonseiMAILab-medllama3-v20}, \textit{BioMedical Llama3 8B} (Bio-Medical-Llama-3-8B)~\cite{ProbeMedicalYonseiMAILab-medllama3-v20}, \textit{BioMistral 7B} (BioMistral-7B)~\cite{labrak2024biomistral}, and \textit{MedAlpaca 7B} (medalpaca-7b) \cite{han2025medalpacaopensourcecollection}. Technical details about the models are in section~\ref{app:references-models}.

\begin{figure}[th!]   
     \centering
      \includegraphics[width=1.0\linewidth]{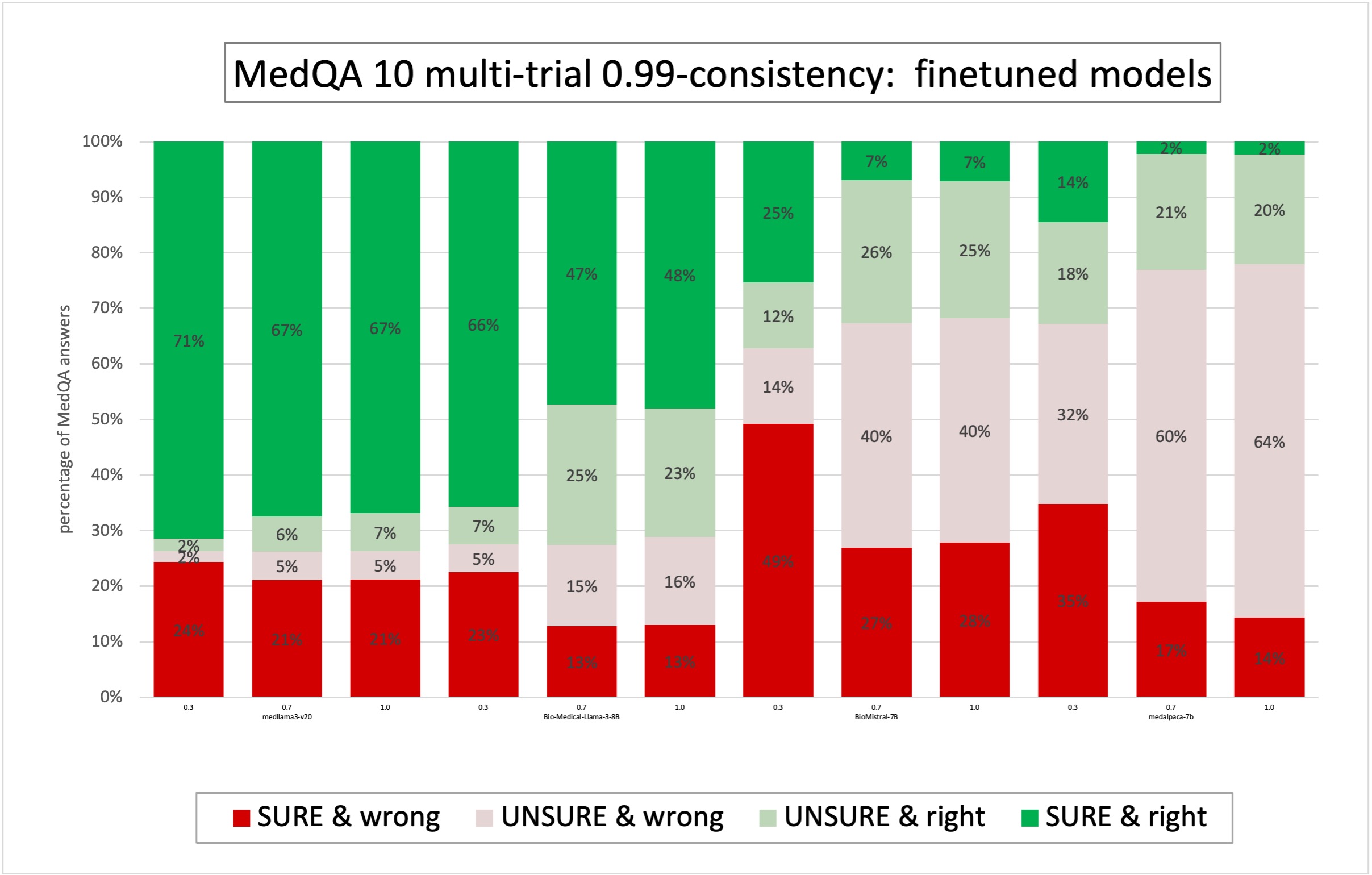}
     \caption{Bar graph showing the proportions of SURE/UNSURE and right/wrong of the 4~finetuned models for the MedQA benchmark, 0.99-consistency, for 3 temperatures.} 
     \label{fig:app-medqa-temperature-finetuned}
\end{figure}

\begin{figure}[th!]   
     \centering
      \includegraphics[width=1.0\linewidth]{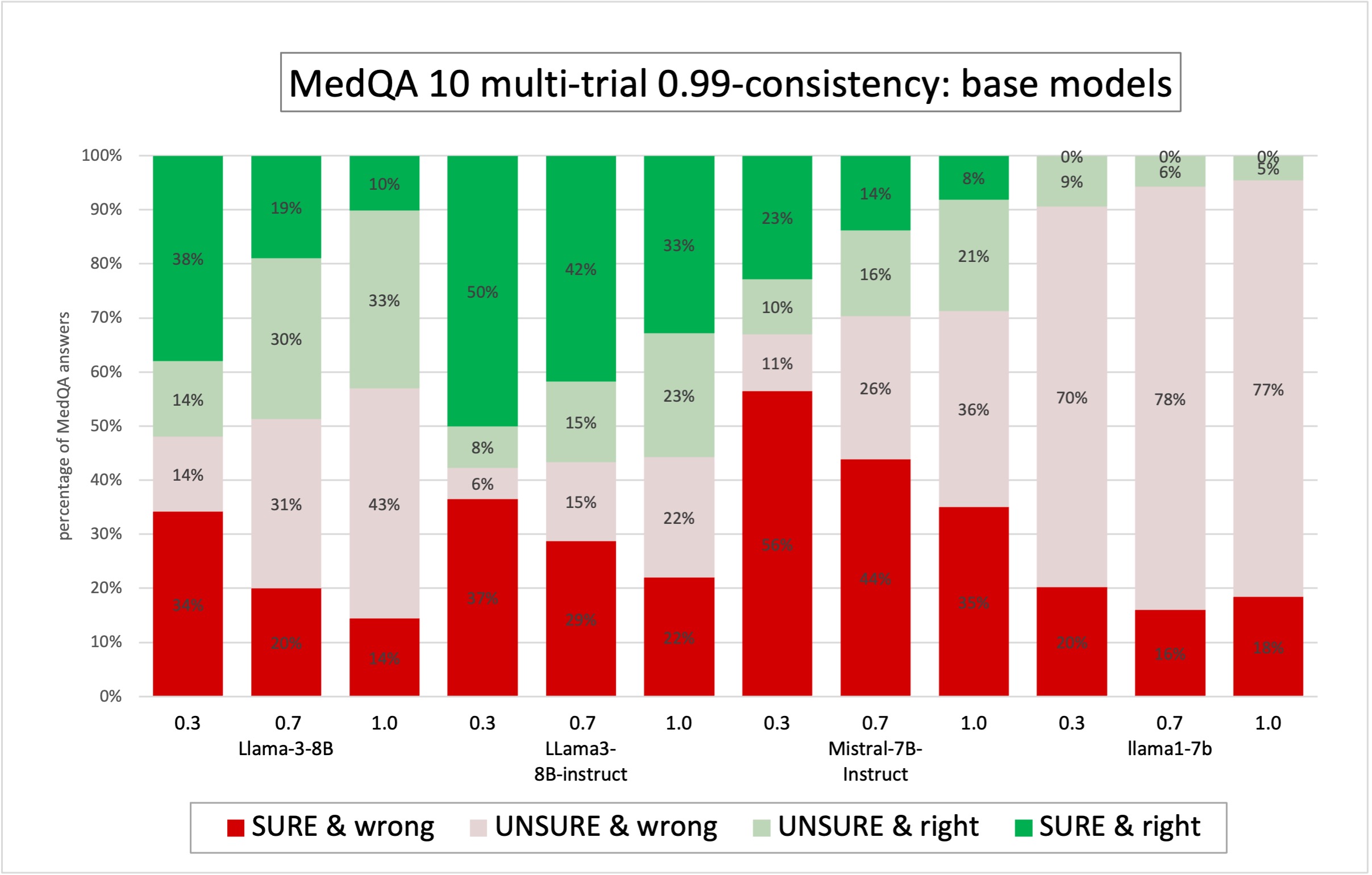}
     \caption{Bar graph showing the proportions of SURE/UNSURE and right/wrong of the 4~base models for the MedQA benchmark, 0.99-consistency, for 3 temperatures.} 
     \label{fig:app-medqa-temperature-base}
\end{figure} 


\begin{figure*}[th!]   
     \centering
      \includegraphics[width=0.8\linewidth]{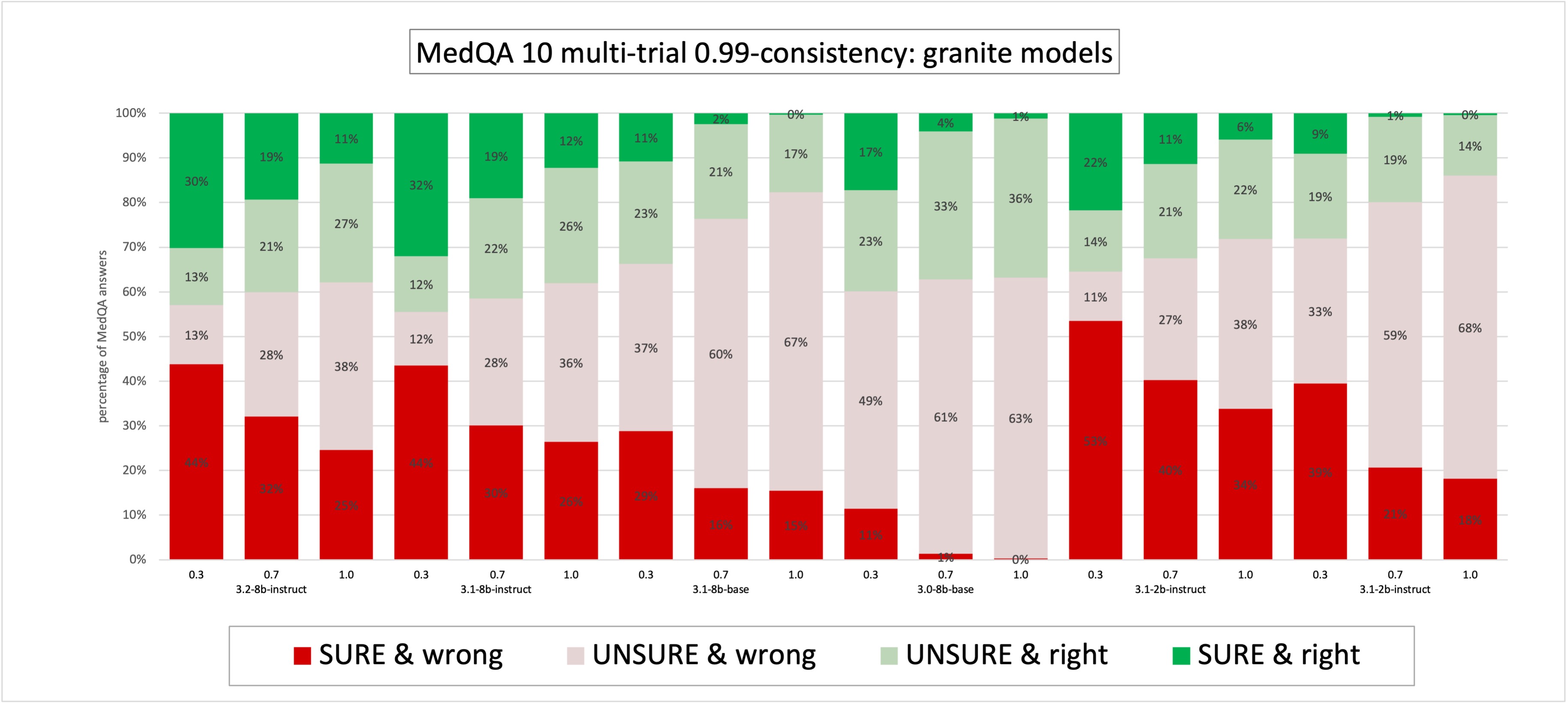}
     \caption{Bar graph showing the proportions of SURE/UNSURE and right/wrong of the 6~granite models for the MedQA benchmark, 0.99-consistency, for 3 temperatures.} 
     \label{fig:app-medqa-temperature-granite}
\end{figure*}

The corresponding 4~base models from where these models were finetuned are: \textit{LLama3 8B} (Llama-3-8B)~\cite{grattafiori2024llama3herdmodels}, \textit{LLama3 8B Instruct} (Llama3-8B-instruct)~\cite{grattafiori2024llama3herdmodels}, \textit{Mistral 7B Instruct} (Mistral-7B-Instruct)~\cite{jiang2023mistral7b}, and \textit{Llama1 7B} (llama1-7b)~\cite{touvron2023llamaopenefficientfoundation}. The granite models were the same used in the MMLU-Redux experiment. 

We considered the exact same infrastructure to run these evaluations, but given the smaller size of MedQA compared to MMLU-Redux, these experiments took only about 1 day of computation time.

\subsection{Finetuned vs. Base Models}

The top of table~\ref{tab:medqa-results} displays the results of the 10~evaluations of the 8~models (finetuned and base) at 3~inference temperatures. Visualizations of the results are provided in figures~\ref{fig:app-medqa-temperature-finetuned} and~\ref{fig:app-medqa-temperature-base}.

First, all the accuracy standard deviations are extremely small, agreeing to the results reported in~\cite{nalbandyan2025score}.
Among the finetuned models, medllama3-v20 had the best accuracy average, $0.736$, at $t=0.3$. The second best model was Bio-Medical-Llama-3-8B, with 0.717 average accuracy, also at 0.3 temperature. 

As for answer consistency, the finetuned medical models produced consistent answers from 49\% to 96\% of the benchmark questions, considering the output at the best temperature but it was as low  as 17\% for the worst model at $t=1$. 
In terms of answer consistency, the best finetuned model was
medllama3-v20 at 0.3 temperature which produced the highest percentage of SURE (S/T) questions, an impressive 96\%, with 0.75 SURE accuracy (RWS), what was higher than its own accuracy average. 



However, the highest ratio of correct SURE questions (RWS) was yielded by the Bio-Medical-Llama-3-8B model at temperatures 0.7 and 1.0, of 0.79, better by more than 5\% over the best average accuracy, $0.736$, although only in about 60\% of the answers. Here is an interesting case where overall accuracy can be increased by filtering out UNSURE answers, albeit at the cost of 40\% of the questions. In some highly critical situations where certainty and correctness are top requirements, this would be the best model. 
The other two models, BioMistral-7B and medalpaca-7b, had much lower performances 
although BioMistral-7B showed a high number of consistent questions but with less than one third correct.




If we look into the S/T column of table~\ref{tab:medqa-results}, we see the same pattern observed in the MMLU-Redux in which, as the temperature increases, the number of SURE questions decreases. However, unlike that case, the number of correct answers among SURE questions (RWS) tends to remain stable or with small decreases, with a few exceptions. 


The middle of table~\ref{tab:medqa-results} displays the results for the 4~base models from where the medical models were finetuned. 
Overall, as expected, the average accuracy is considerably lower. However, the LLama3-8B-instruct model, at 0.7 temperature, had an impressive RWS of 0.58 with 70\% SURE answers, results much better than the worst two finetuned models. Also, the same model at temperature 0.3 yielded a considerable 87\% percentage of SURE questions. 


\subsection{Exploring Multiple Versions of a Model}\label{medqa-granite-results}

We also performed an identical study using the same 6~different versions of the \textit{granite} model~\cite{granite2024granite}. Results are shown on table~\ref{tab:medqa-results} and a visualization of the results are provided in figure~\ref{fig:app-medqa-temperature-granite}. We observed here very similar results as in the previous models: the percentage of SURE questions at the best temperature for each model, going from 24\% to 75\%; the accuracy among SURE questions (RWS) similar to the average accuracy, with some exceptions. However, unlike in the base models, the best RWS performances were at the low temperature 0.3.

\begin{figure}[t!]   
     \centering
      \includegraphics[width=0.9\linewidth]{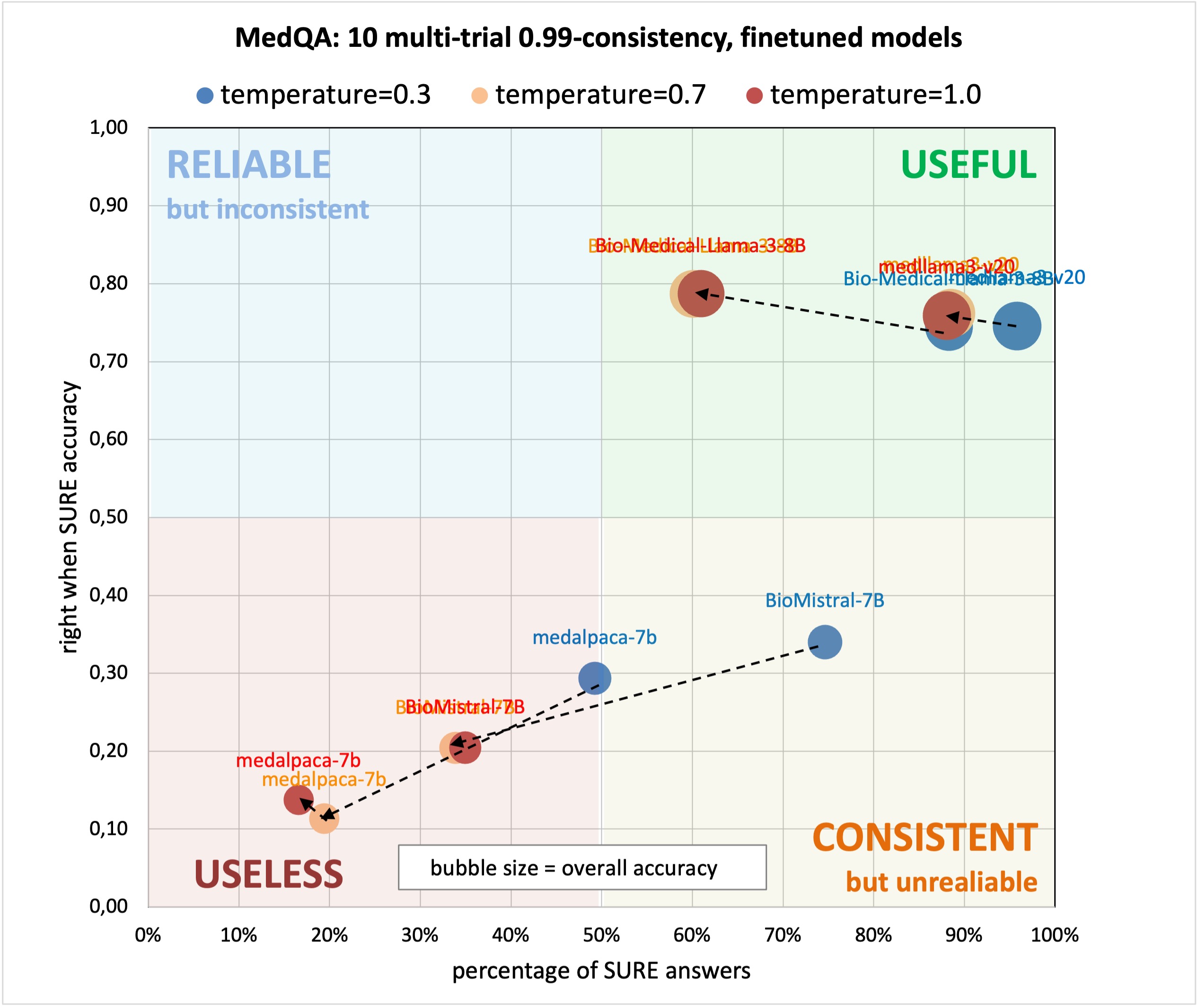} \\
      \vspace{2mm}
      \includegraphics[width=0.9\linewidth]{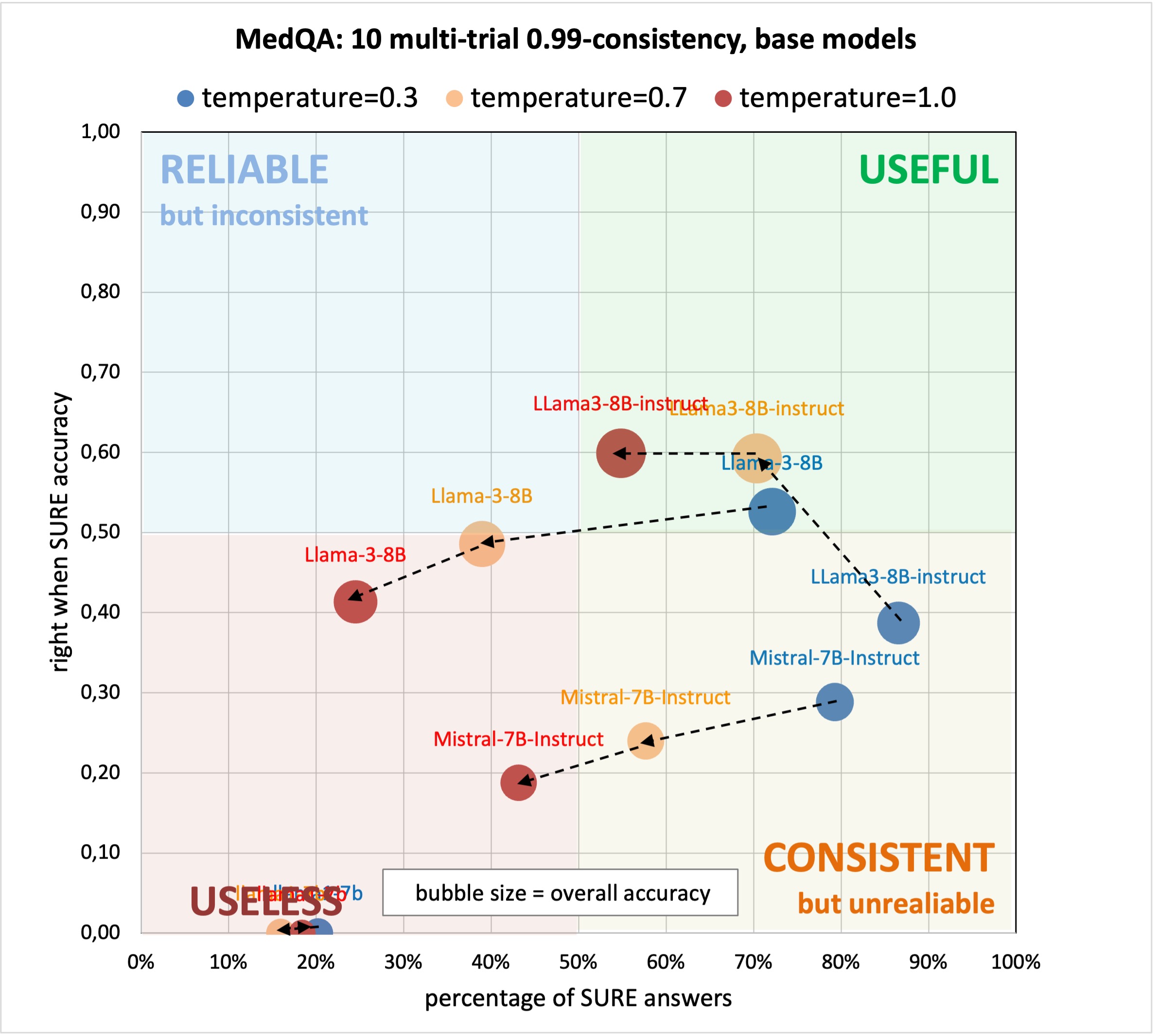} \\
      \vspace{2mm}
      \includegraphics[width=0.9\linewidth]{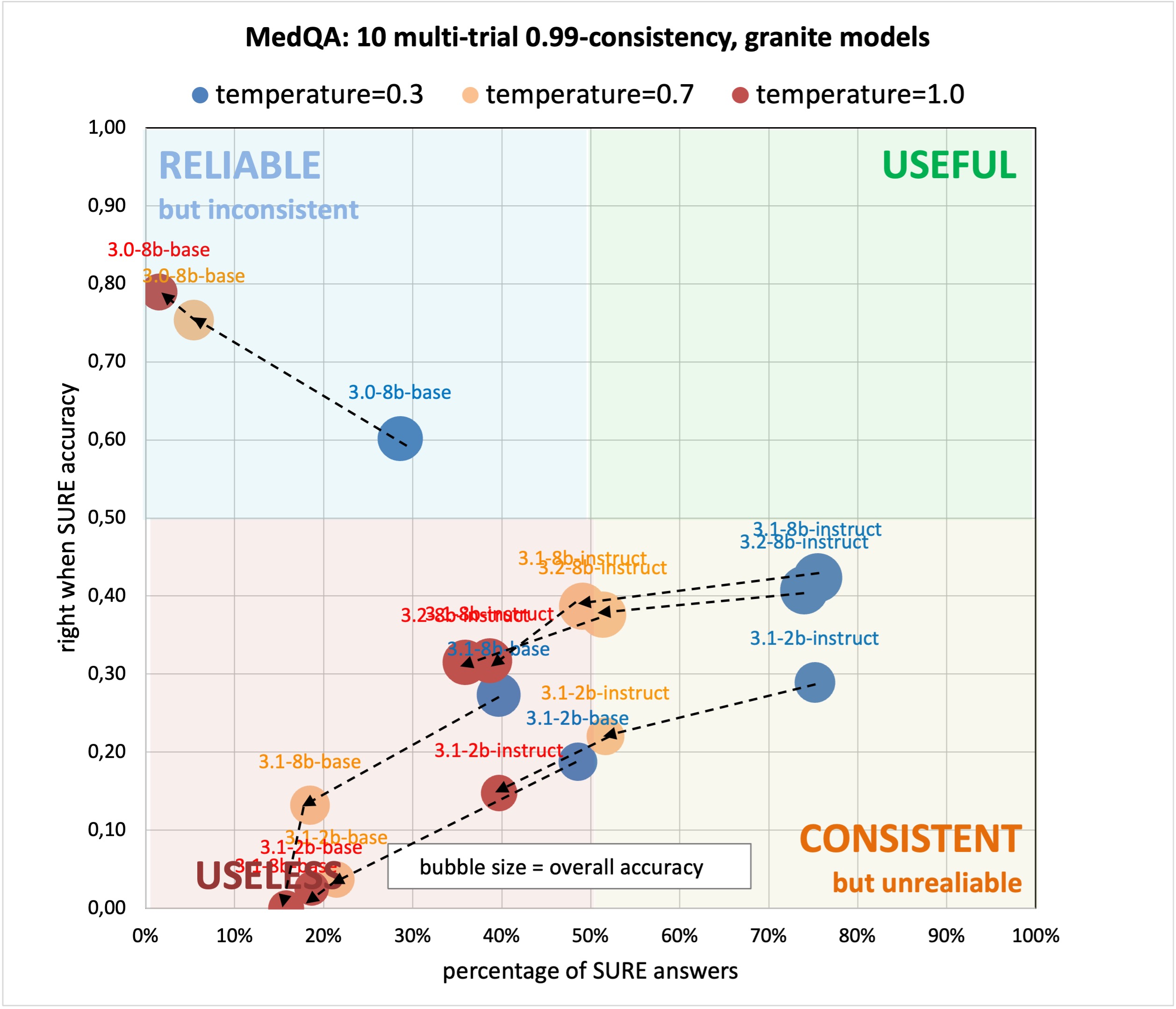}
     \caption{Consistency plots for the MedQA benchmark for finetuned (top), base (center), and granite (bottom) models.} 
     \label{fig:app-medqa-consistency-plots}
\end{figure}

\begin{figure}[th!]   
     \centering
      \includegraphics[width=0.9\linewidth]{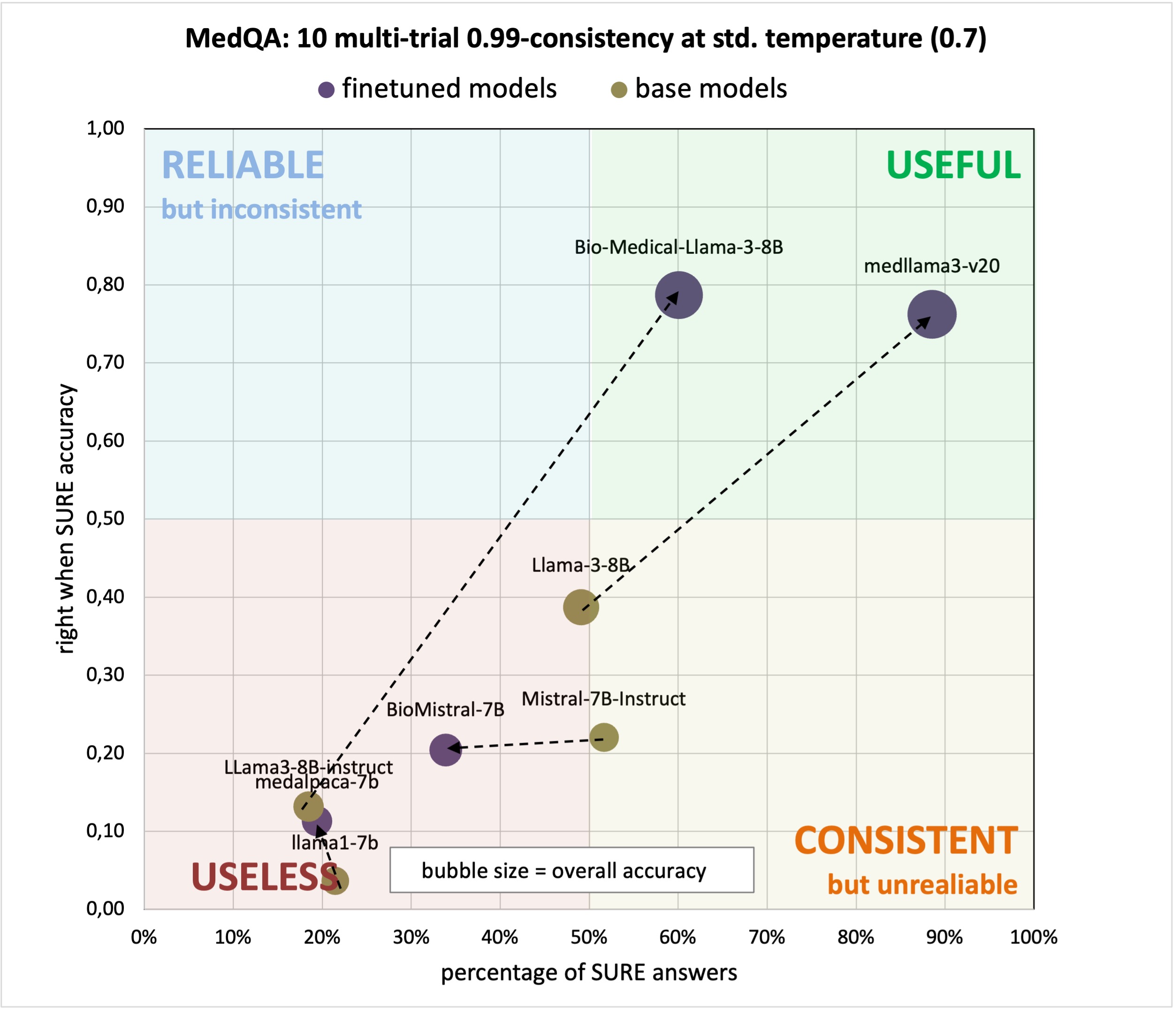} \\
     \caption{Consistency plot for the MedQA benchmark comparing finetuned and base models at standart temperature $t=0.7$.} 
     \label{fig:app-medqa-finetuned-base}
\end{figure}





\subsection{Visualization with  Consistency Plots}

Figure~\ref{fig:app-medqa-consistency-plots} depicts the consistency plots for the finetuned, base, and granite models  based on the results depicted in table~\ref{tab:medqa-results}. As in the case of MMLU-Redux, we see that, as temperature increases, models go through a path from right to left, that is, their inconsistency increases (as measured by the S/T scores) as temperature goes up. However, not all models progress upwards, like in the case of MMLU-Redux, since the accuracy of consistent answers (RWS) only seems to increase for some of the best models. In fact, most of the models go downwards, showing that low temperatures seem to be the best option for MedQA.

Interestingly, as depicted on figure~\ref{fig:app-medqa-finetuned-base}, the two best finetuned medical models progressed upwards left to right when considering the original base and the finetuned versions, showing a clear path of improvement. This was not seen for the two worst models, possibly an indication that the finetuning process was not successful.

\subsection{Aggregated Results over All Models}

\begin{figure}[th!]   
    \centering
     \includegraphics[width=0.95\linewidth]{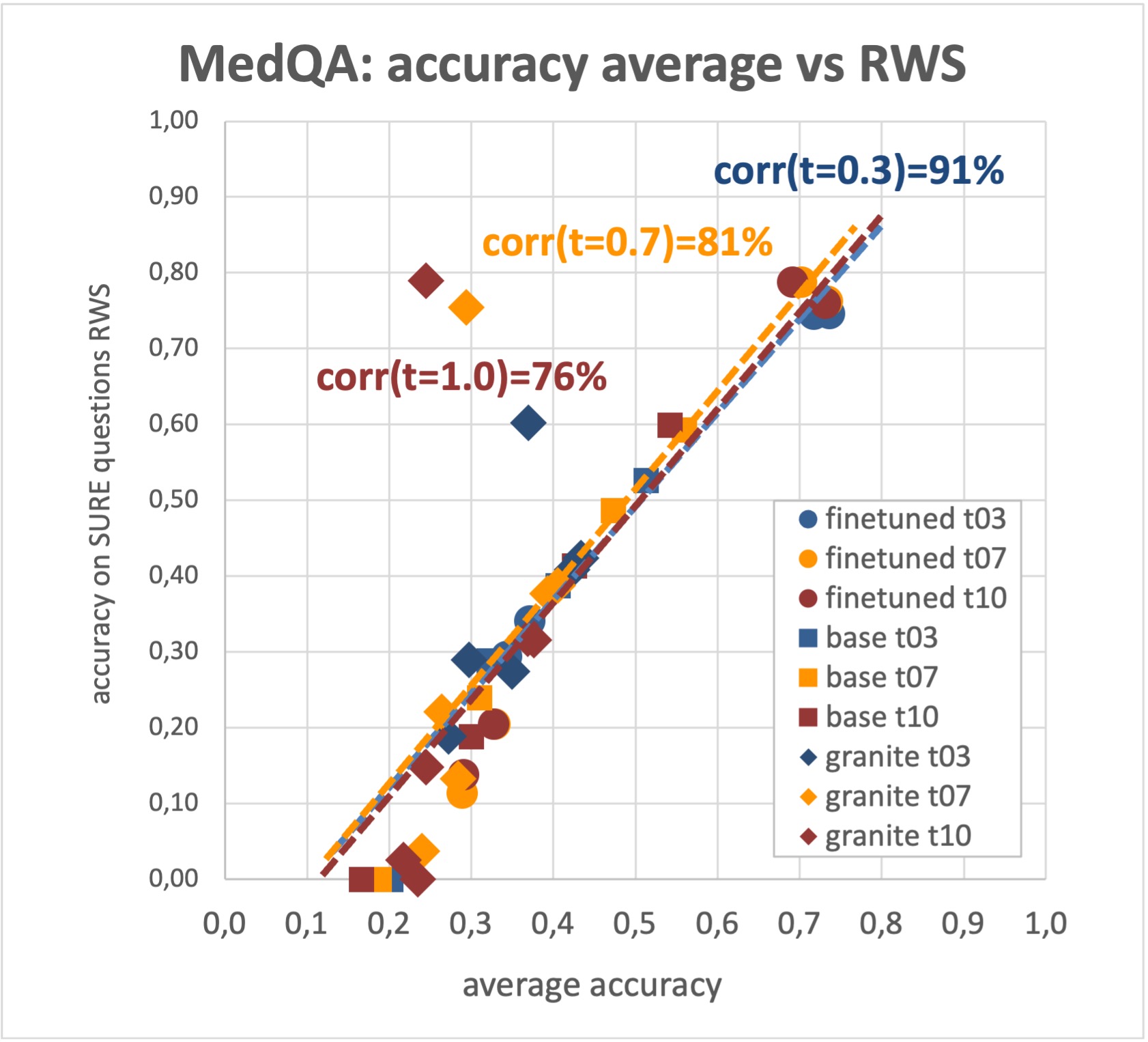} \\
     \vspace{1mm}
     \includegraphics[width=0.95\linewidth]{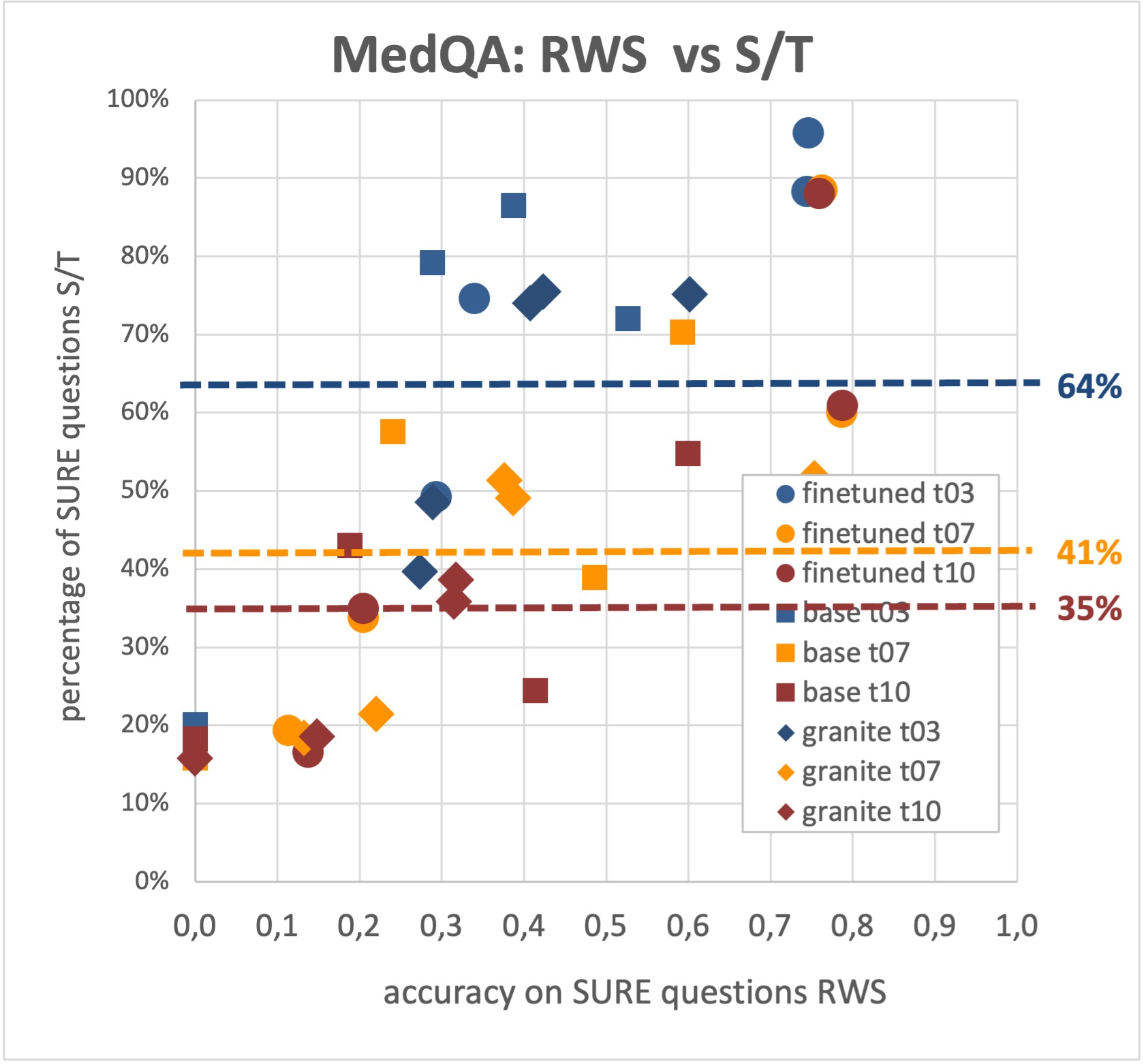}
    \caption{Scatter plot graphs showing the correlation between RWS and average accuracy (left) and between S/T and average accuracy (right) for the 4 finetuned and base models tested with MedQA benchmark across 3 temperatures.} 
    \label{fig:medqa-aggregated}

\end{figure}

As we did with the MMLU-Redux benchmark, we looked into the correlation between the RWS results and the accuracy average. With MedQA we obtained much higher correlation values of 91\%, 81\%, and 76\%, for temperatures 0.3, 0.7, and 1.0, respectively. As seen in the scatter plot depicted on the top of figure~\ref{fig:medqa-aggregated}, except for the 3 granite models (corresponding to the results of 3.0-8b-base), the values of average accuracy and RWS are mostly aligned. Regression for $t=0.3$ values yielded a coefficient of
1.234 and an intercept value of -0.119, with $R^2=0.831$.


Similarly as we did with MMLU-Redux, we computed the percentage of SURE questions (S/T) for each model at the best temperature level and obtained averages of $64\% \pm 14\%$, $42\% \pm 25\%$, and $35\% \pm 23\%$, for temperatures of 0.3, 0.7. an 1.0, respectively. The scatter plot on the bottom of figure~\ref{fig:medqa-aggregated} shows the values for all models and the averages for each temperature. Unlike the case of MMLU-Redux, RWS values occupied a large range from 0\% to 85\%.

\section{Limitations}\label{sec:limitations}

The results and conclusions drawn from these studies may have been impacted by two key limitations. First, 
we are considering only multiple-choice benchmarks in this study, under the top-K sampling decoding method. Although this methodology warrants good precision in the reported results, it may have impacted them, as discussed, for instance, in~\cite{song2024good}. Second, we are not considering possible effects of benchmark contamination~\cite{xu2024benchmark,cheng2025survey,ravaut2024much} which may have inflated the number of consistently corrected answers due to \textit{rogue memorization}~\cite{cavalin2024fixing}, especially for medium models. 

\section{Conclusion and Future Work}\label{sec:conclusion}

This paper proposed a generic definition of answer consistency by an equivalence to oracle guessing at the same level and then derived a formula for answer consistency in the case of repetitions of multiple-choice questions. We then suggested a compact representation  of answer consistency as a pair of numbers, RWS \textbar \hspace{0.3mm} S/T, the proportion of correct answers among the consistent ones (RWS) followed by the percentage of consistent answers (S/T). 
We also proposed a way to visualize this representation, the consistency plot, with which we could more easily understand phenomena like the effects of finetuning and of different inference temperatures. 

We performed experimental studies with the MMLU-Redux and MedQA benchmarks using 26 models at 3 different temperatures. Our results indicated: (i)~most small models produce only between 50\% and 80\% of consistent answers (S/T); (iii)~medium size models display high level of consistency, above 95\%; (iii)~there is a high correlation between model average accuracy and the proportion of correct answers among the consistent ones (RWS) at low temperatures; and (iv)~in the more generic benchmark MMLU-Redux, increasing temperature yielded a higher degree of accuracy among SURE answers, RWS. 

Several aspects remain to be explored. First, we did not consider benchmark contamination through memorization, what may have had significant impacts in the consistency measurements. Although we explored the granite family where the risk of contamination is likely to be smaller, we are still working out reliable ways to filter contaminated questions out. Moreover, we want to go beyond simple repetition and use equivalent wordings for questions and answers, aiming to apply the methods to contexts beyond multiple-choice benchmarks. Finally, we also want to study alternative methods to determine whether a question is consistently answered by a model, during runtime, which are not based on repetitive API calls, a method which is often too expensive to be used in practice.

{
\bibliography{references}
}

\clearpage

\appendix
\textbf{APPENDIXES}

\section{The Models Used in the Experiments}\label{app:references-models}

Table~\ref{tab:references-models-table} provides references and technical details about all the models used in our experiments.

\begin{table*}[ht!]
    \centering
     \resizebox{.9\textwidth}{!}{%
    \scriptsize
    \begin{tabular}{|l|l|l|l|}
        \hline
        \textbf{Model}          & \textbf{Release Date*} & \textbf{Github} & \textbf{Reference} \\
        \hline
        Llama-3-8B              & April 18th, 2024  & \url{ https://github.com/meta-llama/llama-models/blob/main/models/llama3} & ~\cite{grattafiori2024llama3herdmodels} \\
        \hline
        Llama-3-8B-instruct     & April 18th, 2024  & \url{ https://github.com/meta-llama/llama-models/blob/main/models/llama3} & ~\cite{grattafiori2024llama3herdmodels} \\
        \hline
        Deepseek-llm-7b-base    & November 29th, 2023 & \url{ https://github.com/deepseek-ai/DeepSeek-LLM} & ~\cite{deepseekai2024deepseekv3technicalreport} \\
        \hline
        Llama-3.3-70b           & December 6th, 2024  & \url{ https://github.com/meta-llama/llama-models/blob/main/models/llama3\_3} & ~\cite{grattafiori2024llama3herdmodels} \\
        \hline
        Mixtral-8x7b-instruct   & December 9th, 2023 & \url{ https://github.com/mistralai/mistral-inference} & ~\cite{qwen2025qwen25technicalreport} \\
        \hline
        Qwen2-5-72b-instruct    & September 19th, 2024 & \url{ https://github.com/QwenLM/Qwen3} & ~\cite{jiang2024mixtralexperts} \\
        \hline
        Granite-3.2-8b-instruct & February 26th, 2025 & \url{ https://github.com/ibm-granite-community/SageMaker/tree/main/granite-3.2-8b-instruct} & ~\cite{granite2024granite} \\
        \hline
        Granite-3.1-8b-instruct & December 18th, 2024 & \url{ https://github.com/ibm-granite/granite-3.1-language-models} & ~\cite{granite2024granite} \\
        \hline
        Granite-3.1-8b-base     & December 18th, 2024 & \url{ https://github.com/ibm-granite/granite-3.1-language-models} & ~\cite{granite2024granite} \\
        \hline
        Granite-3.0-8b-base     & October 21st, 2024 & \url{ https://github.com/ibm-granite/granite-3.0-language-models} & ~\cite{granite2024granite} \\
        \hline
        Granite-3.1-2b-instruct & December 18th, 2024 & \url{ https://github.com/ibm-granite/granite-3.1-language-models} & ~\cite{granite2024granite} \\
        \hline
        Granite-3.1-2b-base     & December 18th, 2024 & \url{ https://github.com/ibm-granite/granite-3.1-language-models} & ~\cite{granite2024granite} \\
        \hline
        Medllama3-v20           & May 22th, 2024 & \url{ https://huggingface.co/ProbeMedicalYonseiMAILab/medllama3-v20} & ~\cite{ProbeMedicalYonseiMAILab-medllama3-v20} \\
        \hline
        Bio-Medical-Llama-3-8B  & August 8th, 2024 & \url{ https://github.com/zekaouinoureddine/BioMed-LLaMa-3} & ~\cite{ContactDoctor_Bio-Medical-Llama-3-8B} \\
        \hline
        BioMistral-7B           & July 20th, 2024 & \url{ https://github.com/BioMistral/BioMistral} & ~\cite{labrak2024biomistral} \\
        \hline
        Medalpaca-7b            & March 28th, 2023 & \url{ https://github.com/kbressem/medAlpaca} & ~\cite{han2025medalpacaopensourcecollection} \\
        \hline
        Llama1-7b               & February 24th, 2023 & \url{ https://github.com/meta-llama/llama-models/tree/main/models/llama3\_1} & ~\cite{touvron2023llamaopenefficientfoundation} \\
        \hline
        \multicolumn{4}{l}{*When there were no release dates for the model, either the date of the initial Github commit or initial HuggingFace commit were considered as release dates.}
    \end{tabular}%
    }
    \caption{References for each of the models used in experiments.}
    \label{tab:references-models-table}
\end{table*}

\end{document}